\documentclass[11pt]{article}

\usepackage[utf8]{inputenc}
\usepackage{lmodern}     % Latin Modern Type1 fonts (vector, copy/search-safe)
\usepackage[T1]{fontenc}
\usepackage{cmap}        % ToUnicode CMap: searchable/copyable PDF text
\usepackage{amsmath,amssymb}
\usepackage{booktabs}
\usepackage{graphicx}
\usepackage{float}
\usepackage[margin=1in]{geometry}
\usepackage{hyperref}
\usepackage{cleveref}
\usepackage{natbib}
\usepackage{algorithm}
\usepackage{algorithmic}
\usepackage{multirow}
\usepackage{array}
\usepackage{caption}
\usepackage{tikz}
\usetikzlibrary{positioning, arrows.meta, calc, fit}

\hypersetup{colorlinks=true,linkcolor=blue,citecolor=blue,urlcolor=blue}

\newcommand{\doi}[1]{\href{https://doi.org/#1}{\texttt{\detokenize{doi:#1}}}}

\newcommand{\E}{\mathbb{E}}
\newcommand{\pb}{\mathrm{pb}}
\newcommand{\pat}{\mathrm{pat}}
\newcommand{\ci}{\mathrm{ci}}
\newcommand{\pred}{\mathrm{pred}}

\newif\ifanon
\anonfalse

\title{Decision-Path Patterns as Tree Reliability Signals:\\
       Path-based Adaptive Weighting for Random Forest Classification}
\ifanon
  \author{Anonymous Author(s)}
\else
  \author{Youngjoon Park\\Independent Researcher\\\texttt{yj.david.park@gmail.com}}
\fi
\date{}

\begin{document}
\maketitle

% ── Abstract ──────────────────────────────────────────────────────────
\begin{abstract}
Random forests construct each tree with a different, randomised representation of the feature space.
Their uniform voting cannot correct errors in regions where trees with incorrect representations probabilistically outnumber correct ones, even when the ensemble collectively holds enough correct information---a reducible error that this paper addresses.
We propose using the structural pattern of each tree's decision path as an instance-adaptive reliability signal to identify and differentially weight the more reliable trees.
At inference, a random forest reaches its prediction through the root-to-leaf path the sample traverses in each tree, so path-level reliability offers a finer granularity than tree-level weighting can access.
We show that this signal reflects the actual reliability of each tree's decision, and that using it yields a statistically significant accuracy improvement over RF on 36 binary classification benchmarks (Wilcoxon $p < 0.0001$).
Class-recall regression---the typical failure mode of RF correction methods---is measured: zero minority-recall regressions and a single majority-recall regression at the $0.2$\,pp threshold, indicating bias reduction rather than a class trade-off.
We further quantify the reducible error accessible to the method from the fitted RF alone; this estimate correlates strongly with per-dataset gain (Pearson $r = {+}0.840$, $p < 0.0001$).
On the qualifying group it identifies, the method delivers a mean ${+}0.99$\,pp accuracy improvement with strict wins on every dataset (7/0/0); an optional amplification mechanism further raises this to ${+}1.48$\,pp.
\end{abstract}


% ── 1. Introduction ───────────────────────────────────────────────────
\section{Introduction}
\label{sec:intro}

Random forests aggregate an ensemble of decision trees by uniform voting, so that the final prediction at any point reflects all trees' decisions equally \citep{breiman2001}; see \citet{biau2016} for a theoretical overview.
However, because of their randomised building process \citep{breiman1996}, the trees do not carry homogeneous information about the true label at any given point, which leads to uneven reliability of the decision according to the tree and the point.
At some points, trees without proper information may probabilistically outnumber the others; the ensemble then makes a wrong decision as a result, even when it has enough correct information among its members.
This acts as a reducible error of the model, suggesting that each tree's contribution should be evaluated differentially according to its reliability.

To address this, prior work has primarily explored tree-level selection or weighting strategies for ensemble aggregation \citep{dietterich2000}.
WRF measures the reliability of each tree by evaluating its prediction performance on OOB samples and assigns it as a static weight \citep{winham2013}.
Label-conditioned dynamic ensemble selection (e.g., KNORA-Eliminate) uses nearest neighbours of a given query point for evaluation and selects the best trees for each inference \citep{ko2008}.

In this paper, we propose using the pattern of each tree's \emph{Decision Path} as a reliability signal for ensemble aggregation.
At inference each tree contributes only through the single root-to-leaf path the sample traverses, so evaluating the path---rather than the tree as a whole---provides a finer granularity that tree-level approaches cannot access.
Decision paths have been used for post-hoc explanation \citep{lundberg2020} and as input features for refitting \citep{cohenshapira2024pnt}, but not as an instance-adaptive reliability signal for ensemble aggregation.

Specifically, we hypothesise that the sequence of majority-class fluctuations along the decision path carries information about the reliability of the resulting leaf-level prediction.
We confirm that prediction accuracy varies systematically across path patterns, and demonstrate that this signal depends on where the sample lies in feature space and what class the tree predicts---requiring joint conditioning on both axes to be properly isolated.
To preserve the variance-reduction benefit of the full ensemble, we do not alter tree selection but instead differentially weight each tree's contribution.
To turn this signal into an aggregation weight without re-introducing the class bias seen in prior work, we adopt a class-conditional weighting scheme that guarantees $\E[w \mid \pb, \ci] = 1$ by construction.

\paragraph{Contributions.}
\begin{enumerate}
  \item \textbf{Path structure as a reliability signal.} Empirical
  demonstration that the structural pattern of a tree's decision path
  carries information about per-tree prediction reliability, but only
  when interpreted jointly with the tree's predicted class and the
  sample's confidence region---no single pattern dominance generalises
  across these axes.

  \item \textbf{Class confound and its resolution.} Identification of
  a structural class confound in naive path-based weighting (minority-predicting
  trees produce more flip-heavy paths for structural reasons
  unrelated to reliability) and its resolution via a class-conditional
  ratio formulation with $\E[w \mid \pb, \ci] = 1$ by construction.

  \item \textbf{Path-based Adaptive Weighting method.} A concrete
  weighting scheme instantiating the above signal, with a six-type
  flip pattern taxonomy and a per-(region, class) weight table
  estimated via cross-validation.

  \item \textbf{Evaluation on 36 binary classification datasets.}
  Statistically significant accuracy gain over RF (Wilcoxon $p < 0.0001$, $\Delta = +0.0026$)
  with zero minority-recall regressions and a single majority-recall
  regression at the $0.2$\,pp threshold (\Cref{tab:aggregate}),
  robust across forest sizes (100--1\,000 trees);
  the 19 out-of-design-set datasets give $\Delta = +0.0027$ with Wilcoxon $p = 0.0011$ (\Cref{tab:ood}).

  \item \textbf{Applicability prediction from the fitted RF.} Two
  indicators read off the fitted RF's OOB by-products---boundary mass $M$
  and boundary spread $S$---jointly predict the per-dataset gain
  (Pearson $r = {+}0.840$, $p < 0.0001$), allowing the expected
  effectiveness to be estimated before the method is applied; on the
  qualifying group identified this way, the method delivers a mean
  ${+}0.99$\,pp gain with strict wins on every dataset (7/0/0).
  Because $M \cdot S$ estimates the exploitable signal, it can also modulate the weighting strength: amplifying weights in proportion to the estimated reducible error (with the factor selected by cross-validation) raises this to ${+}1.48$\,pp with recall safety unchanged at $K^{*}$.
\end{enumerate}

% ── 2. Related Work ───────────────────────────────────────────────────
\section{Related Work}
\label{sec:related}

\subsection{Path-based interpretation}

Decision paths have been used primarily for post-hoc explanation rather than for ensemble weighting.
TreeSHAP \citep{lundberg2020} computes exact Shapley-value feature attributions by traversing the path each tree takes for a given input, yielding per-feature contributions to a prediction.
Such methods use the path to decompose \emph{what} a tree predicted, not to assess \emph{how reliably} it predicted.
Other work has encoded path information as input features.
\citet{cohenshapira2024pnt} propose PnT, which selects informative paths
from a forest via statistical testing and uses the selected paths as
additional input features for refitting a born-again tree or forest.
The path is treated as a global rule whose general usefulness is
assessed once and then applied uniformly to all samples; this differs
from the present work, where the path's structural pattern serves as a
per-sample reliability signal conditioned on the tree's prediction and
the sample's location.
To our knowledge, the sequential pattern of node-majority-class labels along the decision path has not been used as a reliability signal for ensemble aggregation.

\subsection{Tree weighting in random forests}

A separate line of work modifies the uniform per-tree aggregation of standard RF.
We group these methods by whether the weight depends on the test instance.

\subsubsection{Static (instance-independent) weighting}

\citet{winham2013} proposed weighted Random Forest (WRF), exploring several OOB-based tree-weighting schemes; we adopt $w_t \propto 1/(1 - \mathrm{OOB\_acc}_t)$ (their $1/\mathrm{tPE}_j$ form) as our static-weighting baseline---the simplest non-trivial variant in their set.
Because the weight does not depend on the test instance, it cannot adapt to local boundary structure.

\subsubsection{Instance-adaptive weighting}

A line of instance-adaptive approaches replaces RF's uniform aggregation
with weights or selections that depend on the test instance.
The KNORA family \citep{ko2008}---the standard reference for dynamic
ensemble selection (DES; see \citet{cruz2018} for a survey)---weights or selects trees per test instance
based on nearest-neighbour competence in the training set: KNORA-Eliminate
(KNE) retains only trees that correctly classify all $k$ nearest neighbours
(falling back to a reduced neighbourhood if none qualify), while KNORA-Union
(KNU) weights trees by the number of correctly classified neighbours.
We use both as our instance-adaptive baselines.
Other competence-based DES variants have been proposed \citep{woloszynski2011}, all sharing the instance-adaptive paradigm.
A known concern is that label-conditioned local selection can introduce
minority-recall harm when the selection signal is class-correlated---the
same confound the proposed method resolves by design.

Other instance-adaptive approaches estimate per-tree weights from signals 
other than nearest-neighbour accuracy.
\citet{park2023} introduced \emph{adaptive weighting} for per-sample tree reliability in ensemble aggregation, combining prediction-domain coverage with neighbourhood structure to estimate it post-hoc for RF.
\citet{bertsimas2025} propose Adaptive Forests using mixed-integer
optimisation via the Optimal Predictive-Policy Trees (OP2T) framework,
and \citet{bertsimas_stoumpou2024} propose Enhanced Random Forests,
deriving personalised tree weights from iterative sample-difficulty
estimation.
The present paper differs from all of these in two respects: the weighting
signal is \emph{tree-internal} (extracted from the decision path itself
rather than from external neighbourhood, optimisation, or difficulty
estimates), and the weight formula is \emph{class-conditional}, removing
the class confound that label-conditioned methods inherit by construction.

\subsection{Tree construction diversity}

A complementary line of work improves RF by diversifying the tree construction process itself.
Extremely Randomized Trees \citep{geurts2006} inject additional randomness by randomising split thresholds rather than optimising them.
Double Random Forest \citep{han2020} applies bootstrap resampling at every node during tree construction rather than only at the root, producing larger and more diverse trees.
Heterogeneous Random Forest \citep{kim2024hrf} reduces the reuse of dominant splitting features across trees by down-weighting features that appeared near the root of previously constructed trees, directly targeting ensemble diversity.
These approaches modify \emph{how} trees are built rather than \emph{how} their votes are aggregated; the present work is orthogonal, operating entirely at the aggregation stage on a standard RF.

% ── 3. Framework: Decision-Path Patterns ─────────────────────────────
\section{Framework: Decision-Path Patterns}

This section makes the hypothesis introduced in \Cref{sec:intro} operational by (i)~extracting the path label sequence (\Cref{sec:path}), (ii)~classifying its flip pattern (\Cref{sec:patterns}), and (iii)~defining a region variable that controls for proximity to the forest's decision boundary (\Cref{sec:region}).

\subsection{Decision Path Structure}
\label{sec:path}

For any sample $x$ and tree $t$, let $n_0, n_1, \ldots, n_d$ be the sequence of nodes traversed from root to leaf.
Define the \emph{path label sequence} as $L = (\ell_0, \ell_1, \ldots, \ell_d)$ where $\ell_k = \arg\max_c \mathrm{value}[n_k][c]$, i.e., the majority class at each node.
A \emph{flip} occurs at position $k$ if $\ell_k \neq \ell_{k-1}$.
The pattern of flips---where they occur and how many times the majority class reverses---characterises the path's structural complexity.

Note that the path label sequence is a property of the tree, not the sample: a given leaf always has the same root-to-leaf path.
This enables efficient implementation by pre-computing the pattern for each leaf once per tree (see \Cref{sec:impl}).

\subsection{Flip Pattern Classification}
\label{sec:patterns}

The underlying intuition is that two trees may reach the same leaf prediction yet differ in how they arrived at it: a path whose node-majority class has remained consistent throughout suggests a stable, well-supported decision, whereas a path that has repeatedly changed direction may indicate that the tree struggled to resolve the sample---even if both ultimately predict the same class.
We formalise this by classifying each path into one of six mutually exclusive pattern types based on flip positions and direction reversals, chosen to capture qualitatively distinct flip behaviours (early vs.\ late switching, oscillation, recovery) while keeping each type interpretable and populous enough for stable per-cell estimation in the weight table.
Let $f_1 < f_2 < \cdots < f_k$ be the normalised flip positions ($f \in [0,1]$) and $n_{\mathrm{rev}}$ the number of direction reversals; the thresholds at $\tfrac{1}{3}$ and $\tfrac{2}{3}$ simply partition the normalised path into early, middle, and late thirds.
\Cref{tab:pattern-def} summarises the six pattern types.

\begin{table}[h]
\centering\small
\caption{Flip pattern definitions.}
\label{tab:pattern-def}
\begin{tabular}{@{}ll@{}}
\toprule
Pattern & Condition \\
\midrule
\textsc{noflip}    & no flips ($k = 0$) \\
\textsc{early\_sw} & $k \ge 1$, $f_1 < \tfrac{1}{3}$, $f_k < \tfrac{2}{3}$, $n_{\mathrm{rev}} = 0$ \\
\textsc{late\_sw}  & $k \ge 1$, $f_1 > \tfrac{2}{3}$, $n_{\mathrm{rev}} = 0$ \\
\textsc{oscillat}  & $n_{\mathrm{rev}} \ge 2$, or ($n_{\mathrm{rev}} = 1$ and $f_k \ge \tfrac{2}{3}$) \\
\textsc{recover}   & $n_{\mathrm{rev}} = 1$, $f_k < \tfrac{2}{3}$ \\
\textsc{other}     & $k \ge 1$, $n_{\mathrm{rev}} = 0$, none of the above (e.g.\ $k{=}1$ with $f_1 \in [\tfrac{1}{3}, \tfrac{2}{3}]$) \\
\bottomrule
\end{tabular}
\end{table}

For binary classification (the scope of this paper), consecutive flips always alternate in direction, so $n_{\mathrm{rev}} = \max(0, k-1)$.
This simplifies the classification: paths with $k \ge 3$ flips are always \textsc{oscillat}; $k = 2$ is \textsc{oscillat} or \textsc{recover} depending on the last flip position; $k \le 1$ yields \textsc{noflip}, \textsc{early\_sw}, \textsc{late\_sw}, or \textsc{other}.

\subsection{Region Variable}
\label{sec:region}

Which path pattern should be trusted more may depend on where the sample lies in feature space: near the decision boundary, where the forest is uncertain, the pattern signal is expected to matter most; far from the boundary, where the forest is already confident, the prediction is reliable regardless of path structure.
We confirm this empirically in \Cref{sec:confound}.
In the context of reliability, a sample's proximity to the decision boundary can be measured by the base RF's own prediction confidence \citep{breiman2001}.

We therefore partition the prediction space using \emph{forest-level probability}: for tree $t$, $\mathrm{fp}_t(x) = \hat{P}_{\mathrm{RF}}(x)[\pred_t] \in [0,1]$ is the forest's probability for the class that tree $t$ predicts, bucketed into 10 equal-width intervals of width~0.10.
It measures how strongly the forest backs tree $t$'s call: values near~1 mean the forest agrees with the tree, values near~0.5 place the sample at the decision boundary, and values near~0 mean the forest dissents from tree~$t$. Keyed to the tree's own predicted class, the region refers to the same decision as that class, so the $(\text{region},\text{class})$ conditioning cells stay coherent---a property we exploit in \Cref{sec:confound} to isolate the within-cell pattern signal without an agree/dissent confound.

% ── 4. Diagnostic Evidence ────────────────────────────────────────────
\section{Diagnostic Evidence}
\label{sec:diagnostic}

Throughout this paper we use 36 binary classification benchmarks.
Of these, 17 (the \emph{design set}) were used during method development; the remaining 19 (the \emph{out-of-design set}) were added only after the method was frozen, enabling the overfitting check of \Cref{tab:ood} in \Cref{sec:additional}.
The diagnostics in this section use all 36 datasets unless noted otherwise.

\subsection{Pattern Distribution}

Pattern frequencies vary substantially across datasets, indicating that path structure is dataset-specific.
The mean distribution across 36 datasets is shown in \Cref{tab:pattern-dist}; \textsc{noflip} dominates (54\%) with substantial \textsc{oscillat} (18\%) and \textsc{early\_sw} (13\%) components.

\begin{table}[H]
\centering\small
\caption{Mean pattern frequency across tree--point pairs, 36 datasets.}
\label{tab:pattern-dist}
\begin{tabular}{@{}lr@{}}
\toprule
Pattern & Mean \% \\
\midrule
\textsc{noflip}    & 53.6 \\
\textsc{early\_sw} & 13.2 \\
\textsc{late\_sw}  &  6.9 \\
\textsc{oscillat}  & 17.5 \\
\textsc{recover}   &  4.7 \\
\textsc{other}     &  4.0 \\
\bottomrule
\end{tabular}
\end{table}

\subsection{Raw Accuracy by Pattern}

Aggregated across 36 datasets, pattern types show strongly differentiated accuracy (\Cref{tab:raw-acc}).
The spread is large (\textsc{noflip} 0.9167 vs \textsc{late\_sw} 0.4111) and reproducible across datasets (per-dataset best$-$worst pattern-accuracy gap averaged over the 36 datasets: 0.4674).
However, this raw ranking is misleading: the pooled spread conflates pattern with both predicted class (\Cref{sec:confound}) and confidence region, and once these are controlled for, no single global ranking survives (\Cref{tab:cond-signal,fig:perpoint}). The consequence is empirical: a naive weighting derived from this raw ranking systematically harms accuracy and minority recall (\Cref{sec:naive-ablation}).

\begin{table}[h]
\centering\small
\caption{Pooled pattern accuracy across 36 datasets $\times$ 5 repeats.}
\label{tab:raw-acc}
\begin{tabular}{@{}lrr@{}}
\toprule
Pattern & Acc & vs mean \\
\midrule
\textsc{noflip}    & 0.9167 & $+$0.1046 \\
\textsc{early\_sw} & 0.8762 & $+$0.0642 \\
\textsc{recover}   & 0.8035 & $-$0.0086 \\
\textsc{other}     & 0.7386 & $-$0.0735 \\
\textsc{oscillat}  & 0.6428 & $-$0.1692 \\
\textsc{late\_sw}  & 0.4111 & $-$0.4010 \\
\midrule
\emph{overall}     & \emph{0.8120} & \\
\bottomrule
\end{tabular}
\end{table}

\subsection{Conditioning on Region and Class}
\label{sec:confound}

\begin{figure}[H]
  \centering
  \includegraphics[width=0.75\textwidth]{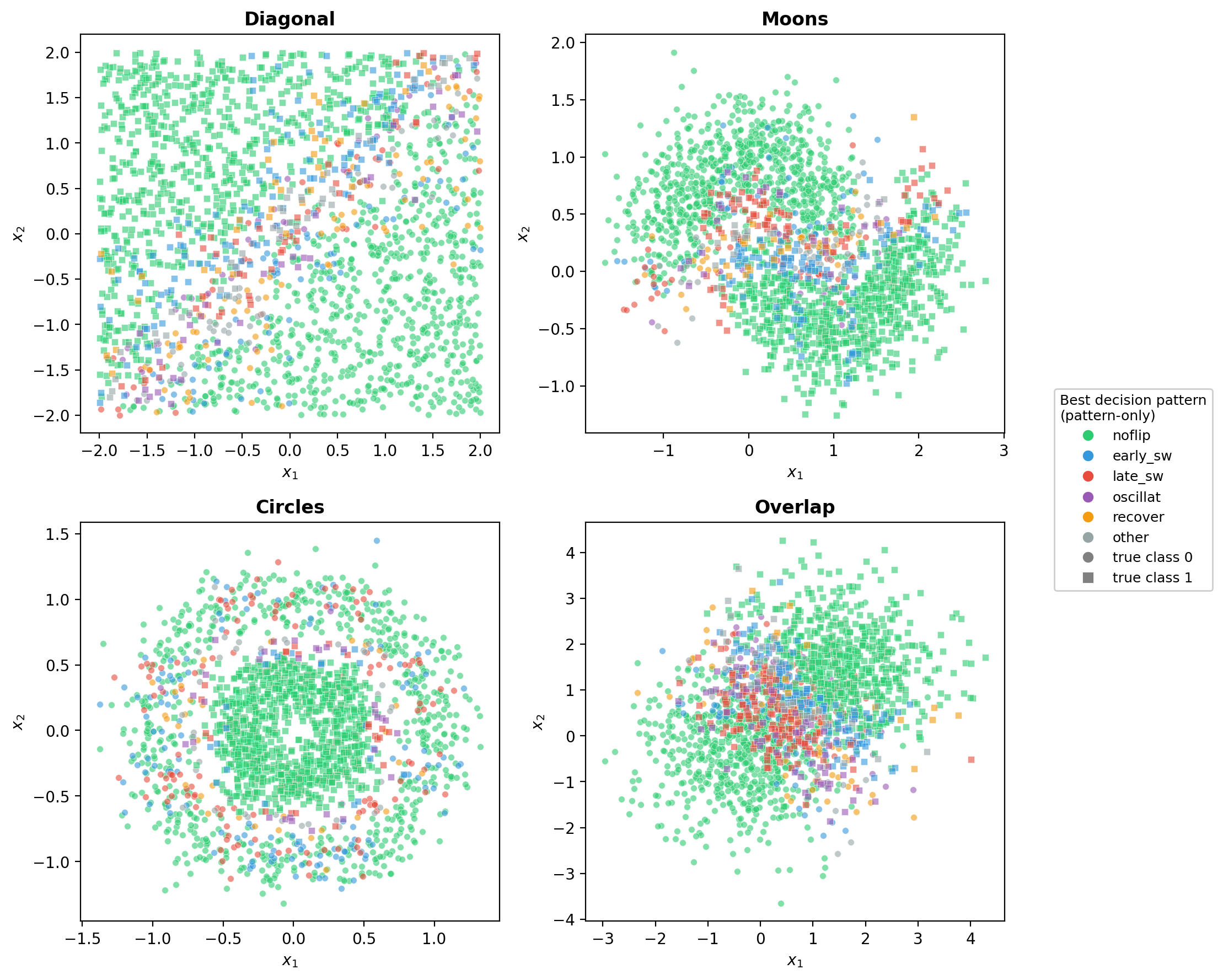}
  \caption{Per-point best decision pattern on four synthetic datasets. Each point is coloured by the flip pattern whose trees are most accurate \emph{on that point}, considering pattern alone---no region or class conditioning; marker shape distinguishes true class ($\circ$ = class 0, $\square$ = class 1). \textsc{noflip} dominates the confident interior while the other patterns concentrate in a band along the decision boundary, consistently across geometries.}
  \label{fig:perpoint}
\end{figure}

\paragraph{Spatial variation.}
The raw accuracy differences in \Cref{tab:raw-acc} raise the question of whether a single pattern ranking generalises across space.
\Cref{fig:perpoint} maps the best pattern at each point on four 2D synthetic datasets with qualitatively different decision boundaries---diagonal (linear), moons (curved), nested circles, and overlapping Gaussians. For each point the best pattern is chosen purely by which flip pattern's trees are most accurate on that point, via 5-fold CV, \emph{without} conditioning on region or class.

The structure is consistent across all four geometries: \textsc{noflip} dominates the confident interior (68--80\% of points) while the remaining patterns concentrate in a band along the decision boundary---the diagonal for Diagonal, the interleaving zone for Moons, the inner/outer ring for Circles, the central overlap for Overlap. Within the boundary band several patterns appear together rather than a single one.

\paragraph{Region: the forest-probability axis.}
This boundary-versus-interior gradient---visible without any region variable---is exactly what a forest's averaged probability captures.
Following \citet{breiman2001}, who frames the random-forest decision as a vote whose averaged outcome approximates the probabilistically most likely class, the per-sample probability $\hat{P}_{\mathrm{RF}}(x)$ tracks the local boundary distance: values near $0.5$ flag samples close to the decision boundary, values near $0$ or $1$ flag confident interiors.
This use of $\hat{P}_{\mathrm{RF}}(x)$ is consistent with random forests producing reasonably well-calibrated probabilities \citep{niculescu2005} and with the forest probability serving as a proxy for prediction uncertainty \citep{mentch2016}.
We therefore adopt the per-tree forest probability $\mathrm{fp}_t(x) = \hat{P}_{\mathrm{RF}}(x)[\pred_t]$ as the region variable (\Cref{sec:region}).

\Cref{tab:region-pat} reports the per-pattern accuracy in each region bucket, pooled across both predicted-class subgroups.
Two observations emerge.
First, every pattern's accuracy is monotone in the region (e.g., \textsc{noflip} rises from $0.0638$ in $[.0,.2)$ to $0.9592$ in $[.8,1.]$; \textsc{late\_sw} from $0.0966$ to $0.9561$), confirming that region captures most of the across-cell accuracy variance.
Second, the within-region spread across patterns is small (max${-}$min within a row: $0.0194$ in $[.8,1.]$ to $0.0411$ in $[.0,.2)$), so region conditioning alone fails to isolate the pattern signal.
We attribute this collapse to a second axis of confounding---predicted class---examined next.

\begin{table}[H]
\centering\small
\caption{Per-pattern accuracy by region, pooled across both predicted-class subgroups (36 datasets $\times$ 5 repeats, 300 trees). Each pattern is monotone in the region (rows reflect forest-backing strength), confirming that region captures most of the across-cell variance. Within-region spread across the six patterns is small (max${-}$min within a row: $0.0194$--$0.0411$), so region alone does not isolate the pattern signal; the class confound (\Cref{tab:confound}) and the conditional analysis (\Cref{tab:cond-signal}) follow.}
\label{tab:region-pat}
\begin{tabular}{@{}lrrrrrr@{}}
\toprule
Region & \textsc{noflip} & \textsc{early\_sw} & \textsc{late\_sw} & \textsc{oscillat} & \textsc{recover} & \textsc{other} \\
\midrule
$[.0,.2)$ & 0.0638 & 0.0555 & 0.0966 & 0.0755 & 0.0607 & 0.0811 \\
$[.2,.4)$ & 0.2432 & 0.2417 & 0.2433 & 0.2547 & 0.2376 & 0.2666 \\
$[.4,.6)$ & 0.5189 & 0.5216 & 0.4930 & 0.5044 & 0.4953 & 0.5297 \\
$[.6,.8)$ & 0.7705 & 0.7724 & 0.7669 & 0.7598 & 0.7576 & 0.7784 \\
$[.8,1.]$ & 0.9592 & 0.9666 & 0.9561 & 0.9472 & 0.9608 & 0.9652 \\
\bottomrule
\end{tabular}
\end{table}

\paragraph{Class confound.}
The within-region spread is small because pattern is confounded with predicted class.
Minority-predicting trees are structurally complex (they partition the feature space more finely to isolate rare-class regions), so reaching a minority leaf typically requires more splits along the path, and the majority class therefore dominates the early-depth nodes before the final minority decision---a late or oscillating flip is the structural consequence, not a reliability deficit.

\begin{table}[H]
\centering\small
\caption{Pattern accuracy by predicted class, 36 datasets, 300 trees. The extreme count asymmetry, in both directions (e.g.\ \textsc{noflip}: 20.6M majority vs 23K minority; \textsc{early\_sw}: 29K majority vs 3.81M minority; \textsc{late\_sw}: 5K majority vs 2.72M minority), confirms the structural origin of the confound.}
\label{tab:confound}
\begin{tabular}{@{}lccr@{}}
\toprule
Pattern & Maj acc ($n$) & Min acc ($n$) & Gap \\
\midrule
\textsc{noflip}    & 0.9167 (20{,}584k) & 0.8805 (23k)    & $+$0.0363 \\
\textsc{early\_sw} & 0.8886 (29k)       & 0.8761 (3{,}809k) & $+$0.0124 \\
\textsc{late\_sw}  & 0.6800 (5k)        & 0.4106 (2{,}724k) & $+$0.2694 \\
\textsc{oscillat}  & 0.6582 (4{,}073k)  & 0.6222 (3{,}022k) & $+$0.0360 \\
\textsc{recover}   & 0.8033 (1{,}898k)  & 0.8294 (10k)      & $-$0.0261 \\
\textsc{other}     & 0.8554 (7k)        & 0.7380 (1{,}245k) & $+$0.1175 \\
\bottomrule
\end{tabular}
\end{table}

\Cref{tab:confound} shows this directly: the count masses for each pattern concentrate on whichever class produces it structurally---\textsc{noflip} on majority (20.6M vs 23K), \textsc{early\_sw}/\textsc{late\_sw} on minority (3.81M vs 29K and 2.72M vs 5K). The large majority$-$minority accuracy gap on \textsc{late\_sw} therefore reflects this count asymmetry rather than a reliability difference between the same trees doing the same work.

\subsection{Conditional Signal after Controlling for Region and Class}

After controlling for region and predicted class, flip pattern type still discriminates accuracy within each cell.
The region variable is the per-tree forest probability $\mathrm{fp}_t(x) = \hat{P}_{\mathrm{RF}}(x)[\pred_t]$ (\Cref{sec:region})---the forest's support for \emph{this tree's own} predicted class---so a cell keyed by (region, predicted class) is automatically a coherent population: the region co-refers with the tree's predicted class, and no cell mixes trees that agree with the forest's vote with trees that dissent from it.
\Cref{tab:cond-signal} reports the within-cell pattern-accuracy spread across the full $[0,1]$ confidence axis.
The spread traces an inverted~U: it peaks in the boundary region $[.4,.6)$---where the forest is split on the tree's call and reweighting is most likely to change the final decision (spread $0.231$/$0.201$ for majority-/minority-predicting trees)---and collapses toward both extremes.
Near $\mathrm{fp}_t{=}1$ the forest strongly backs the tree and the tree is almost always correct (marginal accuracy ${\approx}\,0.96$ for both classes); near $\mathrm{fp}_t{=}0$ the forest strongly dissents and the tree is almost always wrong (marginal accuracy $0.068$ for majority-, $0.091$ for minority-predicting trees); at both extremes the outcome is nearly determined and pattern type adds little.

\begin{table}[h]
\centering\small
\caption{Within-cell pattern-accuracy spread after conditioning on region and predicted (tree) class, 36 datasets $\times$ 5 repeats, 300 trees. The region is the per-tree forest probability $\mathrm{fp}_t(x) = \hat{P}_{\mathrm{RF}}(x)[\pred_t]$ (\Cref{sec:region}); because it co-refers with the tree's own predicted class, each (region, tree class) cell is a coherent population with no agree/dissent confound, so no forest-class axis is needed. Displayed at a coarse 0.20-width bucket resolution for readability ($5\times2=10$ cells); the deployed weight table uses the 0.10-width buckets of \Cref{sec:region}. \emph{Share} is the cell's fraction of all tree--sample pairs across the 36 datasets (totals $\approx$ 37.4\,M pairs); summing the two rows of a region gives the region share. Spread traces an inverted~U across the confidence axis---largest in the boundary region $[.4,.6)$ and small at both extremes, where the marginal accuracy is near~$0$ or near~$1$ and the outcome is nearly determined. All ten cells show spread ${>}\,0.02$.}
\label{tab:cond-signal}
\begin{tabular}{@{}llrrllr@{}}
\toprule
Region & Tree & Share & Spread & Best & Worst & Marg.\ acc \\
\midrule
$[.0,.2)$ & maj & 0.9\%  & 0.030 & \textsc{oscillat} & \textsc{early\_sw} & 0.068 \\
$[.0,.2)$ & min & 3.0\%  & 0.071 & \textsc{late\_sw} & \textsc{noflip}    & 0.091 \\
\addlinespace
$[.2,.4)$ & maj & 2.2\%  & 0.133 & \textsc{oscillat} & \textsc{other}     & 0.246 \\
$[.2,.4)$ & min & 4.2\%  & 0.083 & \textsc{other}    & \textsc{noflip}    & 0.251 \\
\addlinespace
$[.4,.6)$ & maj & 4.3\%  & 0.231 & \textsc{noflip}   & \textsc{late\_sw}  & 0.502 \\
$[.4,.6)$ & min & 4.2\%  & 0.201 & \textsc{noflip}   & \textsc{late\_sw}  & 0.514 \\
\addlinespace
$[.6,.8)$ & maj & 10.4\% & 0.100 & \textsc{other}    & \textsc{oscillat}  & 0.765 \\
$[.6,.8)$ & min & 5.1\%  & 0.119 & \textsc{noflip}   & \textsc{oscillat}  & 0.770 \\
\addlinespace
$[.8,1.]$ & maj & 53.2\% & 0.048 & \textsc{other}    & \textsc{oscillat}  & 0.958 \\
$[.8,1.]$ & min & 12.5\% & 0.023 & \textsc{recover}  & \textsc{oscillat}  & 0.963 \\
\bottomrule
\end{tabular}
\end{table}

The best pattern shifts across confidence regions---for majority-predicting trees it moves from \textsc{oscillat} at low confidence through \textsc{noflip} at the boundary to \textsc{other} at high confidence---confirming that per-cell weight estimation, rather than a global pattern ranking, is necessary.
The near-zero spread at the two extremes does not imply that all patterns are equivalent there; rather, the differences are small enough that the assigned weight has negligible influence on the final prediction once the outcome is already nearly determined.

\subsection{Visualising Conditional Pattern Signal: Region View}

\begin{figure}[H]
  \centering
  \includegraphics[width=\textwidth]{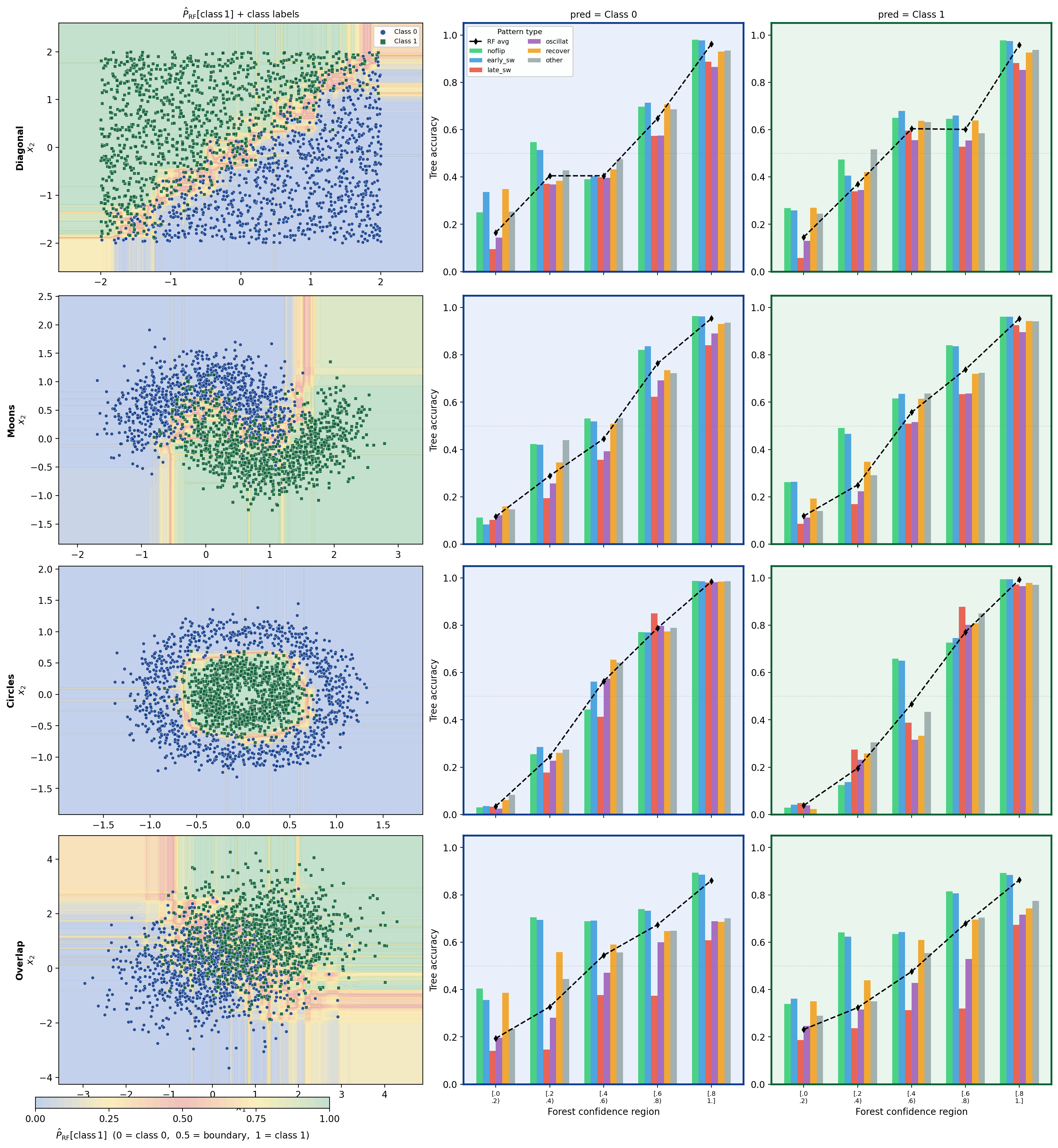}
  \caption{Synthetic 2D experiments. Left: forest probability for class~1, $\hat{P}_{\mathrm{RF}}(x)[\text{class}\,1] \in [0,1]$, with a pastel diverging colormap (light blue\,=\,class~0 region, light green\,=\,class~1 region, light red\,=\,boundary at $0.5$, light yellow\,=\,intermediate); scatter points use deep blue (true class~0) and deep green (true class~1) to match the heatmap regions. Centre and right: per-pattern tree accuracy by region (per-tree forest probability $\mathrm{fp}_t$, $[0,1]$, 5 buckets of 0.20), conditioned on predicted class; subplot border colour identifies the predicted class (blue\,=\,Class~0, green\,=\,Class~1). The dashed line shows RF's marginal accuracy $P(\text{correct} \mid \pb, \ci)$. Note: the heatmap axis is fixed to class~1, while the bar-chart axis $\mathrm{fp}_t$ follows each tree's own predicted class, so for class-0-voting trees the bar-chart bucket corresponds to $1-\hat{P}_{\mathrm{RF}}(x)[\text{class}\,1]$ (mirrored around $0.5$).}
  \label{fig:synthetic}
\end{figure}

The preceding analysis shows that conditioning on (region, class) removes the structural confound and reveals an exploitable signal.
\Cref{fig:synthetic} visualises the resulting per-pattern accuracy curves on the same four synthetic datasets as \Cref{fig:perpoint}, training a 300-tree RF and collecting pattern accuracy via 5-fold CV.

Key observations: (1)~signal concentration---within-cell spread is largest in the mid-confidence regions around the boundary and collapses in the high-confidence region $[.8,1.]$, consistent with the inverted~U of \Cref{tab:cond-signal}; (2)~geometry-dependent pattern ranking---which pattern is most reliable differs across boundary types; (3)~RF baseline crossing---in every dataset, some patterns lie above and others below the RF baseline in the boundary region, suggesting exploitability.

\subsection{Real-Data Diagnostic: Per-Dataset Region-Best}

The preceding diagnostics either pool all 36 datasets (Tables~\ref{tab:raw-acc}--\ref{tab:cond-signal}) or focus on the four synthetic geometries (Figs.~\ref{fig:perpoint}--\ref{fig:synthetic}).
\Cref{tab:region-best} now breaks the real-data signal apart by dataset: for each of the 17 design-set datasets (the only ones consulted during method development; see \Cref{sec:diagnostic}), we report the best pattern in the boundary region $[.4,.6)$ and the adjacent moderate-confidence region $[.6,.8)$, separately for majority- and minority-predicting trees.
The boundary region is where reweighting is most likely to change the final decision (see \Cref{tab:cond-signal}).

\begin{table}[h]
\centering\footnotesize
\caption{Best pattern per dataset in the boundary region and the adjacent moderate-confidence region, 17 design-set datasets (300 trees, 5 repeats, $n \ge 30$ per cell). Four to five distinct best patterns appear in each column---no single ranking applies across datasets.}
\label{tab:region-best}
\setlength{\tabcolsep}{3pt}
\begin{tabular}{@{}l cccc@{}}
\toprule
& \multicolumn{2}{c}{$[.4, .6)$} & \multicolumn{2}{c}{$[.6, .8)$} \\
\cmidrule(lr){2-3} \cmidrule(lr){4-5}
Dataset & maj & min & maj & min \\
\midrule
hepatitis     & oscillat  & early\_sw  & recover   & early\_sw  \\
wine          & late\_sw  & early\_sw  & noflip    & late\_sw   \\
heart-disease & noflip    & early\_sw  & noflip    & early\_sw  \\
parkinsons    & recover   & late\_sw   & recover   & early\_sw  \\
sonar         & recover   & late\_sw   & early\_sw & early\_sw  \\
heart-statlog & early\_sw & recover    & early\_sw & early\_sw  \\
haberman      & noflip    & early\_sw  & noflip    & oscillat   \\
spect         & recover   & early\_sw  & noflip    & early\_sw  \\
ionosphere    & recover   & late\_sw   & oscillat  & early\_sw  \\
australian    & oscillat  & noflip     & early\_sw & early\_sw  \\
breast-w      & noflip    & oscillat   & noflip    & other      \\
transfusion   & noflip    & early\_sw  & noflip    & early\_sw  \\
diabetes      & noflip    & early\_sw  & noflip    & early\_sw  \\
wdbc          & oscillat  & late\_sw   & recover   & late\_sw   \\
banknote      & oscillat  & late\_sw   & oscillat  & other      \\
german-credit & noflip    & early\_sw  & noflip    & late\_sw   \\
tic-tac-toe   & oscillat  & late\_sw   & recover   & early\_sw  \\
\midrule
\# distinct   & 5 & 5 & 4 & 4 \\
\bottomrule
\end{tabular}
\end{table}

Three observations.
First, the best pattern is not universal: four to five distinct patterns appear as best in each column.
Second, majority and minority columns frequently disagree for the same dataset (e.g.\ wine: \textsc{late\_sw} vs \textsc{early\_sw} in $[.4,.6)$), confirming that class conditioning is needed even within a single region.
Third, the same dataset can have different best patterns across regions (e.g.\ australian: \textsc{oscillat} $\to$ \textsc{early\_sw} for majority from $[.4,.6)$ to $[.6,.8)$), confirming that region conditioning is also needed.
These two axes of variation---across datasets and across regions---motivate the per-dataset, per-cell weight table estimated from training data.

% ── 5. Application: Path-based Adaptive Weighting ────────────────────
\section{Application: Path-based Adaptive Weighting}

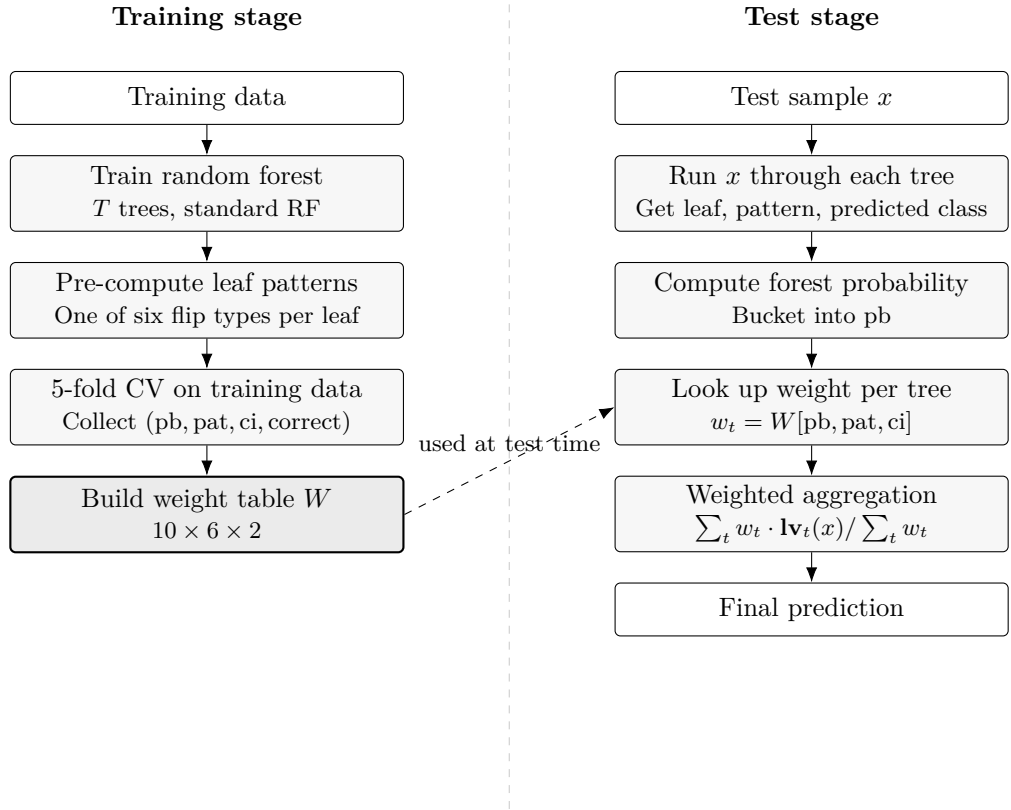
\begin{figure}[H]
\centering
\begin{tikzpicture}[
  node distance=4mm and 12mm,
  font=\small,
  every node/.style={align=center},
  stage/.style={
    rectangle, rounded corners=2pt, draw, thin,
    minimum width=52mm, minimum height=10mm,
    inner sep=2pt, fill=black!3
  },
  io/.style={
    rectangle, rounded corners=2pt, draw, thin,
    minimum width=52mm, minimum height=7mm,
    inner sep=2pt, fill=white
  },
  table/.style={
    rectangle, rounded corners=2pt, draw, thick,
    minimum width=52mm, minimum height=10mm,
    inner sep=2pt, fill=black!8
  },
  arrow/.style={-{Latex[length=2mm]}, thin},
  xfer/.style={-{Latex[length=2mm]}, thin, dashed},
  header/.style={font=\small\bfseries},
  myc/.style={font=\footnotesize},
]

% --- Column headers ---
\node[header] (htrain) at (0, 0) {Training stage};
\node[header] (htest)  at (8, 0) {Test stage};

% --- Vertical divider ---
\draw[gray!40, dashed, thin] (4, 0.4) -- (4, -10.5);

% --- TRAINING column (left) ---
\node[io,   below=of htrain]      (tdata)   {Training data};
\node[stage, below=of tdata]      (forest)  {Train random forest \\ {\footnotesize $T$ trees, standard RF}};
\node[stage, below=of forest]     (pattern) {Pre-compute leaf patterns \\ {\footnotesize One of six flip types per leaf}};
\node[stage, below=of pattern]    (cv)      {5-fold CV on training data \\ {\footnotesize Collect $(\pb, \pat, \ci, \text{correct})$}};
\node[table, below=of cv]         (wtable)  {Build weight table $W$ \\ {\footnotesize $10 \times 6 \times 2$}};

\draw[arrow] (tdata)   -- (forest);
\draw[arrow] (forest)  -- (pattern);
\draw[arrow] (pattern) -- (cv);
\draw[arrow] (cv)      -- (wtable);

% --- TEST column (right) ---
\node[io,   below=of htest]       (tsample) {Test sample $x$};
\node[stage, below=of tsample]    (apply)   {Run $x$ through each tree \\ {\footnotesize Get leaf, pattern, predicted class}};
\node[stage, below=of apply]      (bucket)  {Compute forest probability \\ {\footnotesize Bucket into $\pb$}};
\node[stage, below=of bucket]     (lookup)  {Look up weight per tree \\ {\footnotesize $w_t = W[\pb, \pat, \ci]$}};
\node[stage, below=of lookup]     (agg)     {Weighted aggregation \\ {\footnotesize $\sum_t w_t \cdot \mathbf{lv}_t(x) / \sum_t w_t$}};
\node[io,   below=of agg]         (out)     {Final prediction};

\draw[arrow] (tsample) -- (apply);
\draw[arrow] (apply)   -- (bucket);
\draw[arrow] (bucket)  -- (lookup);
\draw[arrow] (lookup)  -- (agg);
\draw[arrow] (agg)     -- (out);

% --- Cross-column transfer arrow: weight table -> lookup ---
\draw[xfer] (wtable.east) -- node[above, myc, midway] {used at test time} (lookup.west);

\end{tikzpicture}
\caption{Pipeline overview. The training stage builds a weight table $W$ from
the forest's pre-computed leaf patterns (\Cref{sec:path,sec:patterns,sec:impl})
via 5-fold cross-validation (\Cref{sec:weight}). At test time, each tree's
weight is looked up by its (forest-probability bucket, flip pattern, predicted
class) triple, and the forest aggregates with these weights in place of
uniform ones (\Cref{sec:predict}).}
\label{fig:pipeline}
\end{figure}

\Cref{sec:diagnostic} motivates per-cell weighting; this section formalises the resulting scheme.
We replace the uniform per-tree weight $1/T$ in standard RF aggregation,
\begin{equation}
  \hat{P}_{\mathrm{RF}}(c \mid x)
    = \frac{1}{T} \sum_{t=1}^{T}
      \frac{\mathbf{lv}_t(x)[c]}{\|\mathbf{lv}_t(x)\|_1},
  \label{eq:rf}
\end{equation}
with a learned weight $w_t(x)$ derived from tree $t$'s path pattern.
\Cref{fig:pipeline} summarises the training and test pipelines, which the following subsections detail.

\subsection{Conditional Weight Estimation}
\label{sec:weight}

Let $\ci \in \{0{=}\text{majority}, 1{=}\text{minority}\}$ denote the predicted class of tree $t$ for sample $x$.
Define the weight as the ratio of conditional accuracy to marginal accuracy:
\begin{equation}
  w(\pb, \pat, \ci) =
    \frac{P(\text{correct} \mid \pb, \pat, \ci)}
         {P(\text{correct} \mid \pb, \ci)}.
  \label{eq:weight}
\end{equation}

\paragraph{Key property.} $\E[w \mid \pb, \ci] = 1$ by construction---no systematic class bias within any confidence region.
A pattern type that is more accurate than the class average in its region receives $w > 1$ (upweighted); a less accurate pattern receives $w < 1$ (downweighted).

\paragraph{Estimation.}
Given training data, we run 5-fold stratified cross-validation.
On each held-out fold, for every tree--sample pair we record the bucket index $\pb$, pattern $\pat$, predicted class $\ci$, and a correctness indicator. We accumulate two tables: $C[\pb, \pat, \ci]$ = sum of correctness indicators (number correct) and $N[\pb, \pat, \ci]$ = number of pairs in that cell.
The weight table is the cell accuracy normalised by the class-marginal cell accuracy in the same region:
\begin{equation}
  W[\pb, \pat, \ci] =
    \frac{C[\pb,\pat,\ci] \,/\, N[\pb,\pat,\ci]}
         {\sum_{\pat'} C[\pb,\pat',\ci] \,/\, \sum_{\pat'} N[\pb,\pat',\ci]}.
  \label{eq:table}
\end{equation}
Cells with $N[\pb,\pat,\ci] < 30$ fall back to $w = 1.0$ (no adjustment).

\subsection{Weighted Prediction}
\label{sec:predict}

At test time, for each sample $x$:
\begin{equation}
  \hat{P}_{\mathrm{Prop}}(c \mid x) =
    \frac{\sum_t W[\pb_t, \pat_t, \ci_t] \cdot \mathbf{lv}_t(x)[c] / \|\mathbf{lv}_t(x)\|_1}
         {\sum_t W[\pb_t, \pat_t, \ci_t]},
  \label{eq:predict}
\end{equation}
where $\pb_t = \mathrm{bucket}(\mathrm{fp}_t(x))$ is the forest-probability bucket of tree $t$, $\pat_t$ is the pattern type of tree $t$'s leaf for $x$, and $\ci_t$ is the predicted class of tree $t$.

\subsection{Implementation}
\label{sec:impl}

The path label sequence from root to leaf is a fixed property of each leaf---it does not depend on the test sample.
We exploit this by pre-computing the pattern for every leaf in each tree via a single depth-first traversal after training.
At prediction time, only the leaf index is needed; the pattern is then a constant-time array lookup.
This avoids per-sample decision-path traversal, reducing the cost of pattern lookup to $O(1)$.

The weight table $W$ has $10 \times 6 \times 2 = 120$ entries and is stored as a small 3D array.
The total overhead beyond standard RF prediction is: one \texttt{apply()} call per tree, one array lookup per tree, and one weighted sum per sample.

% ── 6. Experimental Evaluation ───────────────────────────────────────
\section{Experimental Evaluation}

The goal of the experiments is to test whether the decision-path reliability signal identified in the preceding sections can improve inference within the standard RF framework; accordingly, the comparators are methods that differentially weight individual trees' predictions based on reliability signals of varying granularity.

\subsection{Datasets}

We evaluate on 36 binary classification benchmark datasets drawn from the UCI \citep{dua2017} and OpenML \citep{vanschoren2014} repositories, which are standard in classifier benchmarking \citep{fernandez2014}; per-dataset characteristics ($n$, $p$, minority class fraction) are listed in \Cref{tab:datasets} in \Cref{app:per-dataset}.
The 17 design-set datasets (hepatitis through tic-tac-toe) are UCI benchmarks used during method development; the 19 out-of-design-set datasets were evaluated only after the method was frozen, providing independent validation (\Cref{tab:ood}).

\subsection{Protocol}

\begin{itemize}
  \item 70/30 stratified train/test split, 30 independent repeats per dataset.
  \item RF: 300 trees; \texttt{max\_features=sqrt}, \texttt{bootstrap=True} (scikit-learn defaults; see \citet{probst2019} for tuning considerations).
  \item Weight table: 5-fold stratified CV on training fold only; test data never used during weight estimation.
  \item Evaluation contract: accuracy primary, with significance assessed by the Wilcoxon signed-rank test \citep{demsar2006} paired at the dataset level over per-dataset mean accuracy across 30 repeats (full-precision sign counts for wins/ties/losses); minority and majority recall must not regress vs RF beyond a $0.2$\,pp practical threshold.
\end{itemize}

\subsection{Comparators}

All methods share the same trained forest (300 trees, \texttt{max\_features=sqrt}, \texttt{bootstrap=True}) and the same train/test split per repeat.

\begin{enumerate}
  \item \textbf{RF}---Standard \texttt{RandomForestClassifier} with uniform aggregation.
  \item \textbf{Proposed}---Path-based Adaptive Weighting (this work). Aggregation weights estimated by 5-fold CV on training data.
  \item \textbf{WRF} \citep{winham2013}---Static tree weighting: $w_t \propto 1/(1 - \mathrm{OOB\_acc}_t)$ (their $1/\mathrm{tPE}_j$ variant).
  \item \textbf{KNORA-E} \citep{ko2008}---Dynamic Ensemble Selection. Eliminates trees that misclassify any of the $k$ nearest training neighbours of the test point. Implementation: \texttt{deslib} 0.3.7 with default $k=7$.
  \item \textbf{KNORA-U} \citep{ko2008}---Dynamic Ensemble Selection. Weights each tree by the count of correctly classified neighbours. Implementation: \texttt{deslib} 0.3.7 with default $k=7$.
\end{enumerate}

This design isolates the aggregation mechanism: RF, the proposed method, and WRF use the same tree predictions with different weights; KNE and KNU select/weight trees per instance based on local competence.

\subsection{Main Results}
\label{sec:aggregate}

\Cref{tab:aggregate} reports the headline comparison of all four weighting schemes against RF on the full 36-dataset benchmark.
The proposed method reaches significance ($\Delta\text{acc} = +0.0026$, $p < 0.0001$, $29/2/5$ wins/ties/losses) with low recall-regression rates on both classes (0/36 minority and 1/36 majority at the $0.2$\,pp threshold); with CV-selected weight amplification (\Cref{sec:amplification}) the mean gain rises to $+0.0036$ with the same win/tie/loss profile and recall-regression counts.
Per-dataset gains vary substantially; \Cref{sec:where-how} analyses when the gain materialises and shows it is predictable from the fitted RF alone.
WRF, KNORA-Eliminate, and KNORA-Union do not reach significance ($p > 0.12$); KNORA-Eliminate has the largest mean minority-recall uplift ($+0.0100$) but pays for it with majority-recall regressions on 15 of 36 datasets and a negative overall accuracy delta.

\begin{table}[H]
\centering\small
\caption{Aggregate comparison vs RF on 36 datasets, all methods sharing the same forest per split. The proposed method shows a strongly significant accuracy gain ($p < 0.0001$). KNE achieves the largest minority-recall uplift but at the cost of accuracy loss and substantial majority-recall instability (15/36 datasets worse by ${>}0.2$\,pp). The proposed method has the lowest recall-regression rates on both classes (0/36 minority and 1/36 majority at the $0.2$\,pp threshold).}
\label{tab:aggregate}
\begin{tabular}{@{}lcccc@{}}
\toprule
& Prop. & WRF & KNE & KNU \\
\midrule
Mean $\Delta$ acc           & $+$0.0026 & $-$0.0010 & $-$0.0023 & $+$0.0002 \\
Wins / Ties / Losses        & 29/2/5    & 19/4/13   & 14/1/21   & 22/1/13   \\
Wilcoxon $p$ (acc)          & \textbf{$<$0.0001} & 0.561 & 0.127 & 0.219 \\
\midrule
Mean $\Delta\, r_{\min}$    & $+$0.0064 & $+$0.0021 & $+$0.0100 & $+$0.0027 \\
$r_{\min}$ worse (${>}0.2$\,pp) & \textbf{0/36} & 2/36 & 5/36 & 3/36 \\
Mean $\Delta\, r_{\mathrm{maj}}$     & $+$0.0012 & $-$0.0019 & \textbf{$-$0.0060} & $-$0.0002 \\
$r_{\mathrm{maj}}$ worse (${>}0.2$\,pp) & \textbf{1/36} & 4/36 & \textbf{15/36} & 4/36 \\
\bottomrule
\end{tabular}
\end{table}

Across all 36 datasets the proposed method matches or exceeds RF accuracy on 31 (29 wins, 2 ties at full precision) and improves minority recall on 28, with no minority-recall regression exceeding $0.2$\,pp and a single majority-recall regression (jm1, $-0.0025$); the largest accuracy regression is $-0.0013$ (ionosphere).
Full per-dataset accuracy and minority recall for all five methods are deferred to \Cref{tab:results} in \Cref{app:results}.

\subsection{Where and How It Works}
\label{sec:where-how}

The mean accuracy gain ($+0.0026$) hides large heterogeneity across the 36 datasets, from $+0.0273$ (mammographic-mass) down to small losses near zero.
This heterogeneity is expected: the reducible error that uniform voting introduces (\Cref{sec:intro}) varies across datasets depending on how much exploitable structure remains in the fitted RF's decisions.
The proposed weight $w = P(\mathrm{correct} \mid \pb,\pat,\ci) / P(\mathrm{correct} \mid \pb,\ci)$ is applied across the full confidence range, but its effect concentrates in the boundary region $[0.4, 0.6)$: in confident regions the pattern-accuracy spread is small (\Cref{tab:region-pat}), so the weight collapses toward~1 and aggregation reverts to uniform voting; in the boundary the spread is largest, and the weight can lift a tree by as much as the pattern-accuracy spread allows.
This suggests that the residual information available in the boundary can be measured from the fitted RF alone, before the method is applied.
Such a measure serves two purposes: it predicts where the method will help, and it opens a route to modulating the method's strength accordingly.

\subsubsection{Measuring the Reducible Error}

We define two per-dataset indicators that estimate the reducible error accessible to the method, both read off the fitted RF's free OOB by-product (\texttt{rf.oob\_decision\_function\_}); no test labels and no weighted prediction are required:
\begin{itemize}
  \item \textbf{Boundary mass} ($M$): fraction of training samples whose OOB-only forest probability for the winning class lies in $[0.4, 0.6)$.
        Measures \emph{how much of the prediction load} the proposed weighting can act on.
  \item \textbf{Boundary spread} ($S$): within the boundary sample set, the max$-$min pattern accuracy across the six flip patterns over the OOB trees, averaged over the two predicted-class subgroups.
        Measures \emph{how much pattern-conditional signal} the boundary actually carries; collapses toward zero when the boundary is noise-dominated.
\end{itemize}
The intuition is that datasets with more boundary samples ($M$) and more pattern-differentiated reliability in that boundary ($S$) leave more room for the method to reduce the uniform-voting error.

\paragraph{Empirical validation.}
\Cref{tab:mass-spread-full} reports $M$, $S$, and the product $M\cdot S$ for every dataset, sorted by $M\cdot S$ (descending).
The product is strongly correlated with $\Delta\text{acc}$ (Pearson $r = +0.840$, $p < 0.0001$; Spearman $\rho = +0.766$, $p < 0.0001$), much stronger than either factor alone ($M$: Pearson $r = +0.138$; $S$: $r = +0.577$).
Univariate $M$ alone or $S$ alone is insufficient---each factor is necessary but neither is sufficient---and the multiplicative interaction is what predicts the gain.

\paragraph{Application prediction.}
The strongest accuracy gains arise where both factors are present.
Datasets that combine non-trivial boundary mass with high boundary spread---mammographic-mass ($M \cdot S = 0.052$, $\Delta\text{acc} = +0.0273$), eeg-eye-state ($0.035$, $+0.0122$), tic-tac-toe ($0.026$, $+0.0095$), and transfusion ($0.023$, $+0.0126$)---account for the largest entries in \Cref{tab:mass-spread-full} and dominate the headline mean.
Datasets that supply only one of the two factors leave less room for the weighting to operate: when mass is small (banknote, musk, oil; $M \le 0.02$) the boundary region itself is too thin for the weight to accumulate effect, and when spread is small (german-credit, spect, ionosphere; $S \le 0.03$) the boundary carries little pattern-conditional signal to extract.
In both bounded regimes the gain stays within $\pm 0.003$.

\begin{table}[H]
\centering\small
\caption{Mean accuracy gain by $M \cdot S$ quintile across the 36 datasets. Each quintile contains 7 datasets (Q1 holds 8 due to $36/5$). Mean $\Delta\text{acc}$ rises monotonically with $M \cdot S$, from near-zero in the bottom quintile to $+0.0099$ in the top quintile, and Q5 has a strict win on every dataset. The product $M \cdot S$ correlates with per-dataset $\Delta\text{acc}$ at Pearson $r = +0.840$ ($p < 0.0001$); per-dataset values are in \Cref{tab:mass-spread-full}.}
\label{tab:mass-spread}
\setlength{\tabcolsep}{6pt}
\begin{tabular}{@{}lcrc@{}}
\toprule
Quintile & $M \cdot S$ range & Mean $\Delta$ acc & W / T / L \\
\midrule
Q1 (lowest)  & [0.0007, 0.0028] & $-$0.0001 & 4 / 1 / 3 \\
Q2           & [0.0031, 0.0046] & $+$0.0006 & 5 / 1 / 1 \\
Q3           & [0.0055, 0.0108] & $+$0.0011 & 7 / 0 / 0 \\
Q4           & [0.0109, 0.0191] & $+$0.0018 & 6 / 0 / 1 \\
Q5 (highest) & [0.0207, 0.0517] & $+$0.0099 & 7 / 0 / 0 \\
\bottomrule
\end{tabular}
\end{table}

\paragraph{Practical use.}
Because $M$ and $S$ are read off the fitted RF alone, they can pre-flag whether the proposed weighting is worth deploying on a new dataset, without running the full weighting pipeline first.
This makes the applicability of the method itself measurable in advance: practitioners can fit RF, compute $M \cdot S$, and decide whether the additional cost of the proposed pipeline is justified by the expected gain on their data.

\subsubsection{Weight Amplification}
\label{sec:amplification}

Beyond predicting where the method works, $M \cdot S$ can also be used to \emph{exploit} that knowledge by modulating the strength of the weight table itself.
Replacing each weight $w$ by $\tilde{w} = \max\!\bigl(1 + \alpha\,(w - 1),\; 0.01\bigr)$ with $\alpha = 1 + K \cdot M \cdot S$ amplifies weight deviations from unity in proportion to the estimated signal strength, while leaving datasets with low $M \cdot S$ essentially unchanged ($\alpha \approx 1$).
The amplification factor $K$ is selected per dataset within the same 5-fold CV that estimates the weight table: for each candidate $K \in \{0, 10, 20, 30\}$, the amplified weights are applied to the CV validation predictions, and the $K$ that maximises CV accuracy is retained.
No test data enter the selection; $K = 0$ corresponds to the unamplified default.
The pseudocode is given in \Cref{alg:amp}.

At the CV-selected $K^{*}$, the mean accuracy gain rises from $+0.0026$ ($K = 0$) to $+0.0036$ against RF, with recall regressions remaining at 0/36 minority and 1/36 majority at the $0.2$\,pp threshold (29/2/5 wins/ties/losses).
The gain is concentrated in the qualifying group: on the top $M \cdot S$ quintile the mean accuracy gain rises from ${+}0.99$\,pp ($K = 0$) to ${+}1.48$\,pp ($K^{*}$), again 7/0/0 vs RF, with the largest single-dataset gain at ${+}3.90$\,pp (mammographic-mass at $K^{*}$).
On the bottom quintile $K^{*} = 0$ in 98\% of runs, leaving those datasets effectively unchanged.
The per-dataset $K^{*}$ sweep is reported in \Cref{tab:amp-sweep-full}.

\subsection{Additional Analyses on Robustness and Ablation}
\label{sec:additional}

This subsection groups four secondary analyses---out-of-design-set validation, dataset-size stratification, tree-count robustness, and a class-conditioning ablation---that test the headline result of \Cref{sec:aggregate} from complementary angles.

\subsubsection{Out-of-Design-Set Validation}

A first concern is whether the gain reflects design-loop overfitting: the 17 design-set datasets were consulted during method development.
Nineteen further datasets were evaluated only after method design was fixed (\Cref{tab:ood}).

\begin{table}[H]
\centering\small
\caption{Results of the proposed method on the 19 out-of-design-set datasets. 15 of 19 are strict wins, 1 tie (oil), 3 losses (bank-marketing, wine-quality-red, thoracic-surgery; all $\le 0.0002$).}
\label{tab:ood}
\begin{tabular}{@{}lrrr@{}}
\toprule
Dataset & $n$ & $\Delta$ acc & $\Delta\, r_{\min}$ \\
\midrule
spambase            &  4\,601 & $+$0.0011 & $+$0.0023 \\
eeg-eye-state       &  4\,000 & $+$0.0122 & $+$0.0183 \\
magic-gamma         &  4\,000 & $+$0.0005 & $+$0.0018 \\
yeast-me2           &  1\,484 & $+$0.0001 & $+$0.0044 \\
oil                 &    937  & $+$0.0000 & $+$0.0111 \\
phoneme             &  5\,404 & $+$0.0007 & $+$0.0001 \\
kr-vs-kp            &  3\,196 & $+$0.0004 & $+$0.0000 \\
jm1                 & 10\,885 & $+$0.0008 & $+$0.0146 \\
electricity         &  4\,999 & $+$0.0002 & $+$0.0013 \\
adult               &  4\,999 & $+$0.0009 & $+$0.0044 \\
musk                &  6\,598 & $+$0.0001 & $+$0.0009 \\
nomao               &  4\,999 & $+$0.0007 & $-$0.0002 \\
bank-marketing      &  4\,999 & $-$0.0001 & $+$0.0015 \\
breast-cancer-uci   &    286  & $+$0.0027 & $+$0.0090 \\
mammographic-mass   &    961  & $+$0.0273 & $+$0.0348 \\
wine-quality-red    &  1\,599 & $-$0.0001 & $+$0.0004 \\
wine-quality-white  &  4\,898 & $+$0.0003 & $+$0.0007 \\
thoracic-surgery    &    470  & $-$0.0002 & $+$0.0016 \\
default-credit-card &  5\,000 & $+$0.0034 & $+$0.0183 \\
\midrule
\textbf{mean}       &         & $+$\textbf{0.0027} & $+$\textbf{0.0066} \\
\emph{W / T / L}    &         & \multicolumn{2}{c}{\emph{15 / 1 / 3}} \\
\emph{Wilcoxon $p$} &         & \multicolumn{2}{c}{\emph{0.0011}} \\
\bottomrule
\end{tabular}
\end{table}

The mean gain on the 19 out-of-design-set datasets is $+0.0027$ with a paired Wilcoxon $p = 0.0011$ (15 strict wins, 1 tie, 3 losses; 16 of 19 match or exceed RF)---comparable to the full 36-dataset mean ($+0.0026$), with the largest contributions from mammographic-mass ($+0.0273$) and eeg-eye-state ($+0.0122$). The effect is clearly not an artefact of design-loop overfitting.

\subsubsection{Performance by Dataset Size}

A second concern is that the proposed method may underperform on small datasets where the per-(region, class, pattern) cells become sparse and the weight table is estimated from few observations.
\Cref{tab:size} tests this by stratifying the proposed method's accuracy gain by dataset size (median split at $n=960$).
The mean gain is comparable across the two size groups ($+0.0023$ small, $+0.0028$ large), with win rates 13/18 and 16/18: the method delivers comparable improvements regardless of dataset size, including the small end ($n \approx 100$--500).

\begin{table}[h]
\centering\small
\caption{Accuracy gain of the proposed method stratified by dataset size (median split at $n=960$). The gain is comparable across the two size groups.}
\label{tab:size}
\begin{tabular}{@{}llrr@{}}
\toprule
Group & $n$ threshold & Mean $\Delta$ acc & Wins \\
\midrule
Small & $\le 960$  & $+$0.0023 & 13/18 \\
Large & $> 960$    & $+$0.0028 & 16/18 \\
\bottomrule
\end{tabular}
\end{table}

\subsubsection{Tree Count Robustness}

A third concern is that the proposed method's advantage could vanish as the forest grows larger, since RF variance decreases with more trees, leaving less room for any reweighting scheme to improve on uniform aggregation.
We test this by sweeping \texttt{n\_estimators} in $\{100, 150, 300, 500, 1000\}$ across all 36 datasets with 30 repeats per configuration (\Cref{tab:trees}).

\begin{table}[H]
\centering\small
\caption{The proposed method vs RF across forest sizes (30 repeats, 36 datasets). $\Delta\text{acc}$ and Wilcoxon $p$ are computed from full-precision per-dataset means; simple subtraction of the displayed RF/Prop. columns may therefore differ from the $\Delta\text{acc}$ column by up to one unit in the last decimal.}
\label{tab:trees}
\begin{tabular}{@{}rcccc@{}}
\toprule
Trees & RF acc & Prop. acc & $\Delta$ acc & Wilcoxon $p$ \\
\midrule
  100 & 0.8691 & 0.8713 & $+$0.0022 & 0.0012 \\
  150 & 0.8693 & 0.8716 & $+$0.0023 & 0.0001 \\
  300 & 0.8699 & 0.8724 & $+$0.0026 & $<$0.0001 \\
  500 & 0.8705 & 0.8726 & $+$0.0021 & 0.0124 \\
1\,000 & 0.8705 & 0.8730 & $+$0.0024 & 0.0004 \\
\bottomrule
\end{tabular}
\end{table}

The proposed method's advantage does not diminish with more trees ($\Delta\text{acc} = +0.0022$ at 100 trees, $+0.0024$ at 1\,000 trees).
This is consistent with the mechanism: more trees improve the weight table estimation (more tree--sample pairs per cell), so the proposed method extracts more of the available signal.
RF accuracy itself saturates between 500 and 1\,000 trees (0.8705 vs 0.8705), but the proposed method continues to benefit from the improved weight estimation.

\subsubsection{Naive Weighting Ablation}
\label{sec:naive-ablation}

Finally, to isolate the role of class conditioning behind the headline gain, we compare two weighting schemes that use the same underlying signal (flip pattern) but differ in whether they condition on predicted class:
\begin{itemize}
  \item \textbf{Naive}: $w = 1 - \text{flip\_rate}$, where $\text{flip\_rate} = k/d$ with $k$ flips over a path of depth $d$ (no class conditioning).
  \item \textbf{Proposed}: $w = P(\text{correct} \mid \pb, \pat, \ci) / P(\text{correct} \mid \pb, \ci)$ (class-conditional ratio).
\end{itemize}

\Cref{tab:naive-ablation} reports the head-to-head against RF under the same 36-dataset, 30-repeat protocol.
Naive weighting trends toward accuracy harm (wins\,=\,17, losses\,=\,18; Wilcoxon $p = 0.819$) and causes minority-recall regression on nearly every dataset (33/36 show any regression; 30/36 exceed 0.2\,pp), while majority recall improves---exactly the one-sided pattern predicted by the class confound (\Cref{sec:confound}).
The proposed class-conditional weighting reverses both pathologies: accuracy becomes significantly positive ($p < 0.0001$), minority-recall regressions (${>}0.2$\,pp) drop from 30 to 0, and majority-recall regressions remain near zero (1/36).

\begin{table}[H]
\centering\small
\caption{Ablation: naive weighting vs class-conditional weighting (proposed), both vs RF. Same 36-dataset, 30-repeat protocol. Without class conditioning, the pattern signal is confounded with predicted class, causing systematic minority-recall harm.}
\label{tab:naive-ablation}
\begin{tabular}{@{}lcc@{}}
\toprule
& Naive & Proposed \\
\midrule
Mean $\Delta$ acc                          & $-$0.0005       & $+$0.0026 \\
Wins / Ties / Losses                       & 17/1/18         & 29/2/5 \\
Wilcoxon $p$ (acc)                         & 0.819           & \textbf{$<$0.0001} \\
\midrule
Mean $\Delta\, r_{\min}$                   & $-$0.0148       & $+$0.0064 \\
$r_{\min}$ worse (${>}0.2$\,pp)            & \textbf{30/36}  & \textbf{0/36} \\
Mean $\Delta\, r_{\mathrm{maj}}$           & $+$0.0050       & $+$0.0012 \\
$r_{\mathrm{maj}}$ worse (${>}0.2$\,pp)    & 1/36            & 1/36 \\
\bottomrule
\end{tabular}
\end{table}

% ── 7. Discussion ─────────────────────────────────────────────────────
\section{Discussion}

\subsection{Effect Size and Conditions for Gain}

The effect of the method is not uniform across datasets but is differentiated by a measurable pre-condition.
\Cref{sec:where-how} identifies two regimes empirically through the boundary mass $M$ and boundary spread $S$, whose product explains most of the per-dataset variance in gain (Pearson $r = {+}0.840$): datasets where RF has left exploitable residual information in the boundary, and datasets where it has not.
In the top $M \cdot S$ quintile, where the residual signal is present, the method delivers a ${+}0.99$\,pp accuracy improvement with strict wins on every dataset (7\,/\,0\,/\,0); with CV-selected weight amplification (\Cref{sec:amplification}) this rises to ${+}1.48$\,pp, again 7/0/0 at $K^{*}$.
In the bottom quintile ($M \cdot S \le 0.0028$), where the signal is largely absent, the gain is near zero (mean ${-}0.0001$, 4 wins / 1 tie / 3 losses), indicating no systematic harm.
The degree of exploitable signal is related to the diversity among trees in the boundary region \citep{kuncheva2003, brown2005}: datasets where trees vote more homogeneously near the boundary leave less room for differential weighting.

The class-conditional ratio weight $w = P(\text{correct} \mid \pb, \pat, \ci) / P(\text{correct} \mid \pb, \ci)$ is a local density ratio in which the $(\pb, \ci)$ conditioning controls for known predictors of accuracy, leaving pattern as the residual signal; when no such residual signal remains, the weight collapses toward 1 and aggregation reverts to RF's behaviour, which is why no systematic harm appears in the low-$M \cdot S$ regime.
The tree-count experiment (\Cref{tab:trees}) supports this interpretation: gains are sustained across forest sizes ($+0.0022$ at 100 trees, $+0.0024$ at 1\,000 trees), and the size stratification (\Cref{tab:size}) shows comparable mean gain across small and large datasets ($+0.0023$ vs $+0.0028$), confirming that signal extraction is stable wherever the signal exists.
Pooled across all 36 datasets the mean gain is $+0.0026$ ($+0.0036$ with CV-selected amplification), though this aggregate mixes the two regimes and is best read alongside the per-quintile breakdown above.

\subsection{Recall Non-regression}
\label{sec:recall-nonregression}

Beyond the conditional effectiveness shown above, the conditional weighting design ($\E[w \mid \pb, \ci] = 1$) also prevents systematic minority-class harm.
At the $0.2$\,pp threshold, the proposed method has the lowest recall-regression rates of any comparator on both classes (0/36 minority, 1/36 majority; \Cref{tab:aggregate}; the single majority regression is jm1 at $-0.0025$), so the accuracy gain operates in the direction of reducing class-conditional errors rather than a one-sided class trade-off. Every other comparator is materially worse, especially on majority recall (WRF 4/36, KNE 15/36, KNU 4/36). All are substantially better than naive weighting (30/36 datasets worse by ${>}0.2$\,pp on minority without class conditioning; \Cref{tab:naive-ablation}).

The comparison with KNE is particularly instructive.
KNE achieves the highest average minority-recall uplift ($+0.0100$) but destabilises both classes: majority recall worsens on 15/36 datasets by ${>}0.2$\,pp, and overall accuracy is negative and non-significant ($\Delta = -0.0023$, $p = 0.127$).
This is consistent with the selection-effect mechanism identified in \Cref{sec:related}: KNE's elimination criterion is label-conditioned (it selects trees that correctly classify neighbours), so when minority neighbours are scarce, the selection signal becomes noisy and harms majority predictions.
The proposed method avoids this by conditioning weights on predicted class rather than on label-conditioned competence.

\subsection{Limitations}
\label{sec:limitations}

Despite the favourable properties above, several caveats qualify the present results.

\paragraph{Researcher degrees of freedom.}
Method design (forest probability as region, 10-bucket resolution, $\mathrm{MIN\_N} = 30$, conditional ratio formula) was iterated with reference to diagnostic results on the 17 design-set datasets.
The reported in-design $p$ should therefore be interpreted alongside the replication on the 19 out-of-design-set datasets (Wilcoxon $p = 0.0011$, \Cref{tab:ood}), which provides cleaner evidence.

\paragraph{Pattern and region discretisation.}
The six-way pattern taxonomy (flip position thresholds at $1/3$ and $2/3$) and the 10-bucket region partition are design choices that prioritise interpretability and stable cell counts for the diagnostic analyses in \Cref{sec:confound}--\Cref{tab:region-best}, not derived from an optimality criterion.
Joint optimisation of taxonomy and bucket resolution could yield stronger predictive gains, at the cost of additional researcher degrees of freedom.

\paragraph{CV overhead.}
Weight table estimation requires five additional RF fits at training time, one per CV fold, on top of the final RF that produces the deployed predictor.
We tested an OOB-based alternative that estimates the weight table from out-of-bag predictions without additional CV fits.
On the 36-dataset benchmark, OOB-based weighting achieves the same mean accuracy gain ($\Delta = +0.0026$, Wilcoxon $p < 0.001$) and zero minority-recall regressions; a direct comparison shows no statistical difference (Wilcoxon $p = 0.937$, per-dataset $\Delta\text{acc}$ correlation $r = 0.990$).
The CV-based estimator is retained as the default because it produces fewer per-dataset accuracy losses (1 vs 5), preserving the near-zero regression property that distinguishes the method from prior work.

\paragraph{Binary classification only.}
The class conditioning ($\ci \in \{0,1\}$) does not directly extend to multi-class settings; conditioning on the tree's predicted class index in $K$-ary classification is a natural extension.

\paragraph{RF variants.}
The proposed method is evaluated on standard RF; how path-pattern distributions and the conditional weight table behave under higher-randomisation variants such as Extremely Randomized Trees \citep{geurts2006} remains an open question.

\subsection{Connection to Prior Work}

We compare the proposed method against the two baseline families on the dimensions where they differ most.

\paragraph{vs WRF (static weighting).}
WRF's global tree weight captures overall tree quality but cannot adapt per instance.
On our 36-dataset benchmark, WRF wins on more datasets than it loses (19 vs 13) but the wins are small while the losses are large (e.g.\ hepatitis $-0.0267$, wine $-0.0162$), yielding a negative mean ($\Delta = -0.0010$, Wilcoxon $p = 0.561$).
The instance-adaptive conditioning on forest probability is the key differentiator: the proposed method's signal is concentrated in the uncertain boundary region where static weights cannot adapt to local boundary structure.
More fundamentally, WRF assigns one weight per tree, whereas the proposed method assigns one weight per tree--sample pair, keyed by the path's flip pattern and the sample's region; since each tree contributes through a single root-to-leaf path at inference, the path is the per-sample active aggregation unit, and path-level weighting accesses refinements that any tree-level scheme---however the per-tree weight is chosen---cannot.

\paragraph{vs KNORA (dynamic selection).}
KNORA-U shows a slightly positive effect ($\Delta = +0.0002$, $p = 0.219$), substantially smaller than the proposed method's gain and accompanied by recall regressions on both classes.
KNORA-E's aggressive elimination creates large per-dataset variance (gains on sonar $+0.0375$ and eeg-eye-state $+0.0223$, losses on transfusion $-0.0327$ and haberman $-0.0316$); the selection-effect mechanism behind this instability is discussed in \Cref{sec:recall-nonregression}.
The proposed method's advantage over both DES methods is that it uses a tree-internal signal (path structure) rather than an external competence measure, and conditions on predicted class to control class bias by construction.

% ── 8. Conclusion ─────────────────────────────────────────────────────
\section{Conclusion}

Random forests treat every tree's vote as equally informative, yet their randomised building process produces trees with uneven reliability across the feature space.
This paper has shown that the structural pattern of each tree's decision path---the sequence of majority-class fluctuations from root to leaf---carries a measurable reliability signal that can be used to differentially weight tree contributions at inference.
The resulting method, Path-based Adaptive Weighting, adopts a class-conditional ratio weight with $\E[w \mid \pb, \ci] = 1$ by construction, preventing the systematic class bias that label-conditioned weighting schemes typically introduce.

On 36 binary classification benchmarks the method achieves a statistically significant accuracy improvement over RF (Wilcoxon $p < 0.0001$; 29 wins, 2 ties, 5 losses; \Cref{tab:aggregate}) with zero minority-recall regressions and a single majority-recall regression at the $0.2$\,pp threshold; this result replicates on the 19 out-of-design-set datasets (Wilcoxon $p = 0.0011$, \Cref{tab:ood}). Under the same protocol, WRF, KNORA-Eliminate, and KNORA-Union all fail to reach significance ($p > 0.12$). The effect is robust across forest sizes from 100 to 1{,}000 trees (\Cref{tab:trees}) and across dataset sizes (\Cref{tab:size}).

The effect of the method is not uniform but conditional, reflecting the varying amount of reducible error across datasets. Two indicators read off the fitted RF's OOB by-products---boundary mass $M$ and boundary spread $S$---jointly quantify this reducible error and predict per-dataset gain (Pearson $r = {+}0.840$): the gain rises monotonically across $M \cdot S$ quintiles, from near zero in the bottom quintile (mean ${-}0.0001$, 4/1/3) to $+0.99$\,pp in the top quintile with strict wins on every dataset (7/0/0); CV-selected weight amplification, which modulates weight strength in proportion to $M \cdot S$ at no additional fitting cost, raises the top-quintile gain to ${+}1.48$\,pp, again 7/0/0 at $K^{*}$.  Where the residual signal is absent, the weight collapses toward 1 and aggregation reverts to RF's behaviour, so the method does not systematically harm performance on datasets where there is nothing to recover. The applicability of the method is thus measurable from the fitted RF before the method itself is applied.

A complementary perspective concerns the level at which aggregation weights are defined. At inference, each tree contributes through a single root-to-leaf path, making the path the per-sample active aggregation unit while trees are the training-time grouping. Tree-level weighting schemes (e.g., WRF) operate at a coarser granularity than this active unit. The present work intervenes at the path-derived level via path-pattern properties, accessing refinements that tree-level approaches cannot.

The structural information encoded in decision paths has been largely overlooked in random forest research. Whether richer path representations or smoother conditional models (isotonic regression, kernel smoothing) can extract more of this signal remains open; the present work establishes that the signal is real, the correction is safe, and its applicability can be measured in advance, with per-dataset gains reaching $+3.90$\,pp (mammographic-mass, $K^{*}$) where the estimated reducible error is largest.

% ── Appendix ──────────────────────────────────────────────────────────
\clearpage
\appendix

\section{Path-based Adaptive Weighting Algorithm}
\label{app:algorithm}

\Cref{alg:paw} gives the end-to-end pseudocode of the proposed method, gathering the training and inference flows of \Cref{fig:pipeline} and Equations~\ref{eq:weight}--\ref{eq:predict} in one place. Pattern classification follows the rules of \Cref{tab:pattern-def} and is applied to any tree's leaf (both the final RF and the inner CV forests use the same function); bucket($\cdot$) maps a forest probability into the 10 equal-width intervals of \Cref{sec:region}.
\Cref{alg:amp} gives the optional weight amplification procedure of \S\ref{sec:amplification}, which modulates weight strength in proportion to $M \cdot S$ with the scaling factor $K^{*}$ selected within the same CV folds; it is applied between weight table construction and inference.

\begin{algorithm}[H]
\caption{Path-based Adaptive Weighting}
\label{alg:paw}
\begin{algorithmic}[1]
\REQUIRE training data $(X_{\mathrm{tr}}, y_{\mathrm{tr}})$, test sample $x$, $T$ trees, MIN\_N $= 30$
\STATE \textbf{Training}
\STATE fit standard RF with $T$ trees on $(X_{\mathrm{tr}}, y_{\mathrm{tr}})$
\FORALL{leaves $\ell$ in every tree $t$}
  \STATE $\mathrm{pat}(\ell) \leftarrow \textsc{ClassifyPattern}(\textsc{Labels}(\ell))$ \COMMENT{\Cref{tab:pattern-def}}
\ENDFOR
\STATE $C[\pb,\pat,\ci] \leftarrow 0$;\quad $N[\pb,\pat,\ci] \leftarrow 0$ \COMMENT{shape $10\times6\times2$}
\FOR{each fold $(X^{(k)}_{\mathrm{tr}}, X^{(k)}_{\mathrm{val}})$ of 5-fold stratified CV on $X_{\mathrm{tr}}$}
  \STATE fit $\mathrm{RF}^{(k)}$ on $X^{(k)}_{\mathrm{tr}}$
  \FORALL{$x' \in X^{(k)}_{\mathrm{val}}$ and each tree $t \in \mathrm{RF}^{(k)}$}
    \STATE $\ell \leftarrow t.\textsc{apply}(x')$;\quad $\ci \leftarrow \arg\max_c t.\mathrm{value}(\ell)[c]$
    \STATE $\pb \leftarrow \textsc{bucket}\!\left(\mathrm{RF}^{(k)}.\textsc{predict\_proba}(x')[\ci]\right)$
    \STATE $N[\pb, \mathrm{pat}(\ell), \ci] \mathrel{+}= 1$;\quad $C[\pb, \mathrm{pat}(\ell), \ci] \mathrel{+}= \mathbf{1}[\ci = y(x')]$
  \ENDFOR
\ENDFOR
\FORALL{cells $(\pb, \pat, \ci)$}
  \IF{$N[\pb, \pat, \ci] \ge \mathrm{MIN\_N}$}
    \STATE $W[\pb,\pat,\ci] \leftarrow \dfrac{C[\pb,\pat,\ci]/N[\pb,\pat,\ci]}{\sum_{\pat'} C[\pb,\pat',\ci]\,/\,\sum_{\pat'} N[\pb,\pat',\ci]}$
  \ELSE
    \STATE $W[\pb,\pat,\ci] \leftarrow 1$ \COMMENT{fallback for sparse cells}
  \ENDIF
\ENDFOR
\STATE \textit{(optionally apply \Cref{alg:amp} to $W$ before inference)}
\STATE \textbf{Inference for sample $x$}
\STATE $\mathrm{fp}(x) \leftarrow \mathrm{RF}.\textsc{predict\_proba}(x)$
\STATE $\boldsymbol{\mathrm{psum}} \leftarrow \mathbf{0}$;\quad $\mathrm{wsum} \leftarrow 0$
\FORALL{trees $t \in \mathrm{RF}$}
  \STATE $\ell \leftarrow t.\textsc{apply}(x)$;\quad $\ci \leftarrow \arg\max_c t.\mathrm{value}(\ell)[c]$
  \STATE $\pb \leftarrow \textsc{bucket}(\mathrm{fp}(x)[\ci])$
  \STATE $w \leftarrow W[\pb, \mathrm{pat}(\ell), \ci]$
  \STATE $\boldsymbol{\mathrm{psum}} \mathrel{+}= w \cdot t.\mathrm{value}(\ell)\,/\,\|t.\mathrm{value}(\ell)\|_1$
  \STATE $\mathrm{wsum} \mathrel{+}= w$
\ENDFOR
\RETURN $\arg\max_c \boldsymbol{\mathrm{psum}}[c] / \mathrm{wsum}$
\end{algorithmic}
\end{algorithm}

\begin{algorithm}[H]
\caption{Weight Amplification via $M \cdot S$}
\label{alg:amp}
\begin{algorithmic}[1]
\REQUIRE weight table $W$ from \Cref{alg:paw}, fitted RF, CV fold predictions, $K$-candidates $= \{0, 10, 20, 30\}$
\STATE compute boundary mass $M$ from RF's OOB by-products \COMMENT{\Cref{sec:where-how}}
\STATE compute boundary spread $S$ from RF's OOB by-products
\FOR{each $K \in K$-candidates}
  \STATE $\alpha \leftarrow 1 + K \cdot M \cdot S$
  \FORALL{cells $(\pb, \pat, \ci)$}
    \STATE $\tilde{W}[\pb,\pat,\ci] \leftarrow \max\!\bigl(1 + \alpha\,(W[\pb,\pat,\ci] - 1),\; 0.01\bigr)$
  \ENDFOR
  \STATE evaluate accuracy on CV validation folds using $\tilde{W}$
\ENDFOR
\STATE $K^{*} \leftarrow \arg\max_{K} \text{CV-accuracy}$
\STATE $\alpha^{*} \leftarrow 1 + K^{*} \cdot M \cdot S$
\FORALL{cells $(\pb, \pat, \ci)$}
  \STATE $W[\pb,\pat,\ci] \leftarrow \max\!\bigl(1 + \alpha^{*}\,(W[\pb,\pat,\ci] - 1),\; 0.01\bigr)$
\ENDFOR
\RETURN amplified $W$
\end{algorithmic}
\end{algorithm}

\section{Per-Dataset Experimental Details}
\label{app:per-dataset}

This appendix groups the per-dataset material that supports the main-text claims: characteristics of the 36 benchmark datasets, the full region-best-pattern table for the 17 design-set datasets, the per-dataset size-effect breakdown, the boundary mass and spread underlying \Cref{sec:where-how}, and the full five-method per-dataset comparison.

\subsection{Dataset Characteristics}
\label{app:datasets}

\begin{table}[H]
\centering\small
\caption{Dataset characteristics ($n$ = preprocessed sample count). Datasets below the rule are in the out-of-design set.}
\label{tab:datasets}
\begin{tabular}{@{}lrrr@{}}
\toprule
Dataset & $n$ & $p$ & Minority \% \\
\midrule
hepatitis      &     83 &  19 & 20.6 \\
wine           &    130 &  13 & 33.1 \\
heart-disease  &    214 &  13 & 45.9 \\
parkinsons     &    195 &  22 & 24.6 \\
sonar          &    208 &  60 & 46.6 \\
heart-statlog  &    270 &  13 & 44.4 \\
haberman       &    306 &   3 & 26.5 \\
spect          &    267 &  22 & 20.6 \\
ionosphere     &    351 &  34 & 35.9 \\
australian     &    690 &  14 & 44.5 \\
breast-w       &    683 &   9 & 34.5 \\
transfusion    &    748 &   4 & 23.8 \\
diabetes       &    768 &   8 & 34.9 \\
wdbc           &    569 &  30 & 37.3 \\
banknote       &  1\,372 &   4 & 44.4 \\
german-credit  &  1\,000 &  20 & 30.0 \\
tic-tac-toe    &    958 &   9 & 34.7 \\
\midrule
breast-cancer-uci    &    286 &   9 & 29.7 \\
thoracic-surgery     &    470 &  16 & 14.9 \\
oil                  &    937 &  49 &  4.3 \\
mammographic-mass    &    961 &   5 & 46.3 \\
yeast-me2            &  1\,484 &   8 &  2.1 \\
wine-quality-red     &  1\,599 &  11 & 46.5 \\
kr-vs-kp             &  3\,196 &  36 & 47.8 \\
eeg-eye-state        &  4\,000 &  14 & 44.8 \\
magic-gamma          &  4\,000 &  10 & 35.2 \\
spambase             &  4\,601 &  57 & 39.4 \\
wine-quality-white   &  4\,898 &  11 & 33.5 \\
electricity          &  4\,999 &   8 & 42.4 \\
adult                &  4\,999 &  14 & 23.9 \\
nomao                &  4\,999 & 118 & 28.6 \\
bank-marketing       &  4\,999 &  16 & 11.7 \\
default-credit-card  &  5\,000 &  23 & 22.1 \\
phoneme              &  5\,404 &   5 & 29.3 \\
musk                 &  6\,598 & 167 & 15.4 \\
jm1                  & 10\,885 &  21 & 19.3 \\
\bottomrule
\end{tabular}
\end{table}

\subsection{Region-Best Pattern (Full Table)}
\label{app:region-best}

\Cref{tab:region-best-full} extends \Cref{tab:region-best} to all five confidence regions for both majority- and minority-predicting trees across the 17 design-set datasets.

Two observations that the main-text \Cref{tab:region-best} cannot show.
First, the diversity of best patterns persists across all five regions---even in the high-confidence region $[.8,1.]$, four (majority) and three (minority) distinct patterns appear as best, suggesting that residual signal exists even where the within-cell accuracy spread is small (\Cref{tab:cond-signal}).
Second, the across-region variation is itself dataset-specific: some datasets keep a single best pattern across all five regions (haberman, transfusion, and diabetes for majority-predicting trees---all \textsc{noflip}), while others change pattern several times.
Both observations support the per-(region, class) cell design of the weight table.

\begin{table}[H]
\centering\scriptsize
\caption{Best pattern per dataset across all five confidence regions (17 design-set datasets, 300 trees, 5 repeats, $n \ge 30$ per cell). ``--'' indicates insufficient data in the cell.}
\label{tab:region-best-full}
\setlength{\tabcolsep}{2.5pt}
\begin{tabular}{@{}l ccccc ccccc@{}}
\toprule
& \multicolumn{5}{c}{Majority-predicting trees} & \multicolumn{5}{c}{Minority-predicting trees} \\
\cmidrule(lr){2-6} \cmidrule(lr){7-11}
Dataset & {[.0,.2)} & {[.2,.4)} & {[.4,.6)} & {[.6,.8)} & {[.8,1.]} & {[.0,.2)} & {[.2,.4)} & {[.4,.6)} & {[.6,.8)} & {[.8,1.]} \\
\midrule
hepatitis     & nf & nf & os & rc & nf  & es & ls & es & es & es \\
wine          & nf & os & ls & nf & nf  & nf & nf & es & ls & nf \\
heart-disease & nf & rc & nf & nf & rc  & ot & es & es & es & es \\
parkinsons    & nf & os & rc & rc & nf  & es & ls & ls & es & es \\
sonar         & nf & rc & rc & es & nf  & nf & rc & ls & es & nf \\
heart-statlog & rc & ot & es & es & ls  & es & nf & rc & es & nf \\
haberman      & nf & nf & nf & nf & nf  & es & es & es & os & es \\
spect         & -- & nf & rc & nf & nf  & es & es & es & es & es \\
ionosphere    & nf & nf & rc & os & rc  & ls & os & ls & es & es \\
australian    & nf & nf & os & es & ot  & ot & os & nf & es & es \\
breast-w      & rc & os & nf & nf & nf  & ls & es & os & ot & es \\
transfusion   & nf & nf & nf & nf & nf  & es & es & es & es & ot \\
diabetes      & nf & nf & nf & nf & nf  & es & es & es & es & es \\
wdbc          & nf & rc & os & rc & nf  & ls & ls & ls & ls & es \\
banknote      & nf & os & os & os & nf  & es & ls & ls & ot & es \\
german-credit & os & rc & nf & nf & nf  & es & es & es & ls & es \\
tic-tac-toe   & nf & nf & os & rc & nf  & es & ls & ls & es & es \\
\midrule
\# distinct   & 3 & 4 & 5 & 4 & 4 & 4 & 5 & 5 & 4 & 3 \\
\bottomrule
\end{tabular}
\\[4pt]
{\footnotesize nf = \textsc{noflip}, es = \textsc{early\_sw}, ls = \textsc{late\_sw}, os = \textsc{oscillat}, rc = \textsc{recover}, ot = \textsc{other}.}
\end{table}

\subsection{Per-Dataset Size Effect (Full Table)}
\label{app:size-effect}

\Cref{tab:size-full} reports the per-dataset accuracy and minority-recall difference between the proposed method and RF (30 repeats), stratified by dataset size.
Note the heterogeneity within each size group: the small group includes both substantial gains (transfusion $+0.0126$) and small losses (spect $-0.0004$, ionosphere $-0.0013$); the large group spans a wide range from negligible (musk $+0.0001$) to substantial (mammographic-mass $+0.0273$, eeg-eye-state $+0.0122$, tic-tac-toe $+0.0095$).
The comparable mean across size groups (\Cref{tab:size}) reflects central tendency---per-dataset variance remains larger in the small group.

\begin{table}[H]
\centering\small
\caption{Per-dataset accuracy and minority-recall difference (Proposed $-$ RF), 30 repeats, augmented with the boundary mass $M$, boundary spread $S$, and product $M \cdot S$ from \Cref{sec:where-how}. Datasets are grouped by the median-split threshold ($n = 960$).}
\label{tab:size-full}
\setlength{\tabcolsep}{4pt}
\begin{tabular}{@{}lrrrrrr@{}}
\toprule
Dataset & $n$ & $M$ & $S$ & $M \cdot S$ & $\Delta$ acc & $\Delta\, r_{\min}$ \\
\midrule
\multicolumn{7}{@{}l}{\emph{Small ($n \le 960$)}} \\
\midrule
hepatitis            &     83 & 0.118 & 0.056 & 0.0066 & $+$0.0027 & $+$0.0067 \\
wine                 &    130 & 0.023 & 0.179 & 0.0041 & $+$0.0000 & $+$0.0000 \\
parkinsons           &    195 & 0.083 & 0.184 & 0.0152 & $+$0.0040 & $+$0.0133 \\
sonar                &    208 & 0.272 & 0.114 & 0.0309 & $+$0.0026 & $+$0.0023 \\
heart-disease        &    214 & 0.144 & 0.075 & 0.0109 & $+$0.0036 & $+$0.0146 \\
spect                &    267 & 0.072 & 0.022 & 0.0016 & $-$0.0004 & $+$0.0000 \\
heart-statlog        &    270 & 0.153 & 0.148 & 0.0228 & $+$0.0016 & $+$0.0000 \\
breast-cancer-uci    &    286 & 0.162 & 0.115 & 0.0187 & $+$0.0027 & $+$0.0090 \\
haberman             &    306 & 0.152 & 0.126 & 0.0191 & $+$0.0007 & $+$0.0069 \\
ionosphere           &    351 & 0.046 & 0.016 & 0.0007 & $-$0.0013 & $-$0.0009 \\
thoracic-surgery     &    470 & 0.045 & 0.076 & 0.0034 & $-$0.0002 & $+$0.0016 \\
wdbc                 &    569 & 0.038 & 0.155 & 0.0060 & $+$0.0023 & $+$0.0031 \\
breast-w             &    683 & 0.021 & 0.083 & 0.0018 & $+$0.0003 & $+$0.0014 \\
australian           &    690 & 0.084 & 0.070 & 0.0058 & $+$0.0002 & $+$0.0025 \\
transfusion          &    748 & 0.091 & 0.253 & 0.0230 & $+$0.0126 & $+$0.0253 \\
diabetes             &    768 & 0.184 & 0.058 & 0.0108 & $+$0.0013 & $+$0.0045 \\
oil                  &    937 & 0.018 & 0.098 & 0.0017 & $+$0.0000 & $+$0.0111 \\
tic-tac-toe          &    958 & 0.106 & 0.246 & 0.0262 & $+$0.0095 & $+$0.0230 \\
\cmidrule{2-7}
\emph{mean}          &         &       &       &        & $+$\emph{0.0023} & $+$\emph{0.0069} \\
\emph{wins}          &         &       &       &        & \multicolumn{2}{c}{\emph{13\,/\,18}} \\
\midrule
\multicolumn{7}{@{}l}{\emph{Large ($n > 960$)}} \\
\midrule
mammographic-mass    &    961 & 0.089 & 0.582 & 0.0517 & $+$0.0273 & $+$0.0348 \\
german-credit        & 1\,000 & 0.228 & 0.024 & 0.0055 & $+$0.0001 & $-$0.0011 \\
banknote             & 1\,372 & 0.006 & 0.549 & 0.0035 & $+$0.0021 & $+$0.0022 \\
yeast-me2            & 1\,484 & 0.011 & 0.105 & 0.0012 & $+$0.0001 & $+$0.0044 \\
wine-quality-red     & 1\,599 & 0.159 & 0.085 & 0.0134 & $-$0.0001 & $+$0.0004 \\
kr-vs-kp             & 3\,196 & 0.015 & 0.305 & 0.0046 & $+$0.0004 & $+$0.0000 \\
eeg-eye-state        & 4\,000 & 0.176 & 0.197 & 0.0346 & $+$0.0122 & $+$0.0183 \\
magic-gamma          & 4\,000 & 0.093 & 0.049 & 0.0045 & $+$0.0005 & $+$0.0018 \\
spambase             & 4\,601 & 0.040 & 0.079 & 0.0031 & $+$0.0011 & $+$0.0023 \\
wine-quality-white   & 4\,898 & 0.132 & 0.051 & 0.0067 & $+$0.0003 & $+$0.0007 \\
electricity          & 4\,999 & 0.132 & 0.033 & 0.0044 & $+$0.0002 & $+$0.0013 \\
adult                & 4\,999 & 0.110 & 0.109 & 0.0119 & $+$0.0009 & $+$0.0044 \\
nomao                & 4\,999 & 0.041 & 0.060 & 0.0025 & $+$0.0007 & $-$0.0002 \\
bank-marketing       & 4\,999 & 0.070 & 0.040 & 0.0028 & $-$0.0001 & $+$0.0015 \\
default-credit-card  & 5\,000 & 0.098 & 0.211 & 0.0207 & $+$0.0034 & $+$0.0183 \\
phoneme              & 5\,404 & 0.081 & 0.122 & 0.0098 & $+$0.0007 & $+$0.0001 \\
musk                 & 6\,598 & 0.004 & 0.516 & 0.0020 & $+$0.0001 & $+$0.0009 \\
jm1                  & 10\,885 & 0.096 & 0.116 & 0.0111 & $+$0.0008 & $+$0.0146 \\
\cmidrule{2-7}
\emph{mean}          &         &       &       &        & $+$\emph{0.0028} & $+$\emph{0.0058} \\
\emph{wins}          &         &       &       &        & \multicolumn{2}{c}{\emph{16\,/\,18}} \\
\bottomrule
\end{tabular}
\end{table}

\subsection{Per-Dataset Boundary Mass and Spread (Full Table)}
\label{app:mass-spread}

\Cref{tab:mass-spread-full} reports the per-dataset boundary mass $M$, boundary spread $S$, and product $M \cdot S$ that underlie the quintile summary of \Cref{tab:mass-spread}, sorted by $M \cdot S$ (descending). Both quantities are read off the fitted RF's OOB by-products; no test labels or weighted prediction are required. The accompanying $\Delta\text{acc}$ column is the Proposed $-$ RF accuracy difference (cf.\ \Cref{tab:results}).

\begin{table}[H]
\centering\small
\caption{Per-dataset boundary mass $M$ (sample-level OOB forest probability $\in [0.4, 0.6)$), boundary spread $S$ (max$-$min pattern accuracy within the boundary region, averaged over the two predicted-class subgroups; OOB-only), their product $M \cdot S$, and the resulting $\Delta\text{acc}$ (Proposed $-$ RF). Sorted by $M \cdot S$ descending. Note: $M \cdot S$ is computed from the full-precision values of $M$ and $S$ before rounding, so the displayed three-decimal $M$ and $S$ may not reproduce the displayed $M \cdot S$ exactly in the last digit.}
\label{tab:mass-spread-full}
\setlength{\tabcolsep}{5pt}
\begin{tabular}{@{}lrrrr@{}}
\toprule
Dataset & $M$ & $S$ & $M \cdot S$ & $\Delta$ acc \\
\midrule
mammographic-mass    & 0.089 & 0.582 & 0.0517 & $+$0.0273 \\
eeg-eye-state        & 0.176 & 0.197 & 0.0346 & $+$0.0122 \\
sonar                & 0.272 & 0.114 & 0.0309 & $+$0.0026 \\
tic-tac-toe          & 0.106 & 0.246 & 0.0262 & $+$0.0095 \\
transfusion          & 0.091 & 0.253 & 0.0230 & $+$0.0126 \\
heart-statlog        & 0.153 & 0.148 & 0.0228 & $+$0.0016 \\
default-credit-card  & 0.098 & 0.211 & 0.0207 & $+$0.0034 \\
haberman             & 0.152 & 0.126 & 0.0191 & $+$0.0007 \\
breast-cancer-uci    & 0.162 & 0.115 & 0.0187 & $+$0.0027 \\
parkinsons           & 0.083 & 0.184 & 0.0152 & $+$0.0040 \\
wine-quality-red     & 0.159 & 0.085 & 0.0134 & $-$0.0001 \\
adult                & 0.110 & 0.109 & 0.0119 & $+$0.0009 \\
jm1                  & 0.096 & 0.116 & 0.0111 & $+$0.0008 \\
heart-disease        & 0.144 & 0.075 & 0.0109 & $+$0.0036 \\
diabetes             & 0.184 & 0.058 & 0.0108 & $+$0.0013 \\
phoneme              & 0.081 & 0.122 & 0.0098 & $+$0.0007 \\
wine-quality-white   & 0.132 & 0.051 & 0.0067 & $+$0.0003 \\
hepatitis            & 0.118 & 0.056 & 0.0066 & $+$0.0027 \\
wdbc                 & 0.038 & 0.155 & 0.0060 & $+$0.0023 \\
australian           & 0.084 & 0.070 & 0.0058 & $+$0.0002 \\
german-credit        & 0.228 & 0.024 & 0.0055 & $+$0.0001 \\
kr-vs-kp             & 0.015 & 0.305 & 0.0046 & $+$0.0004 \\
magic-gamma          & 0.093 & 0.049 & 0.0045 & $+$0.0005 \\
electricity          & 0.132 & 0.033 & 0.0044 & $+$0.0002 \\
wine                 & 0.023 & 0.179 & 0.0041 & $+$0.0000 \\
banknote             & 0.006 & 0.549 & 0.0035 & $+$0.0021 \\
thoracic-surgery     & 0.045 & 0.076 & 0.0034 & $-$0.0002 \\
spambase             & 0.040 & 0.079 & 0.0031 & $+$0.0011 \\
bank-marketing       & 0.070 & 0.040 & 0.0028 & $-$0.0001 \\
nomao                & 0.041 & 0.060 & 0.0025 & $+$0.0007 \\
musk                 & 0.004 & 0.516 & 0.0020 & $+$0.0001 \\
breast-w             & 0.021 & 0.083 & 0.0018 & $+$0.0003 \\
oil                  & 0.018 & 0.098 & 0.0017 & $+$0.0000 \\
spect                & 0.072 & 0.022 & 0.0016 & $-$0.0004 \\
yeast-me2            & 0.011 & 0.105 & 0.0012 & $+$0.0001 \\
ionosphere           & 0.046 & 0.016 & 0.0007 & $-$0.0013 \\
\bottomrule
\end{tabular}
\end{table}

\subsection{Per-Dataset Method Comparison (Full Table)}
\label{app:results}

\begin{table}[H]
\centering\footnotesize
\caption{Per-dataset accuracy and minority recall ($r_{\min}$), 30 repeats each. All methods share the same forest per split. Wins/ties/losses (\Cref{tab:aggregate}) are computed at full precision and may differ from the 3-decimal display here.}
\label{tab:results}
\setlength{\tabcolsep}{3.5pt}
\begin{tabular}{@{}lr rrrrr rrrrr@{}}
\toprule
& & \multicolumn{5}{c}{Accuracy} & \multicolumn{5}{c}{Minority Recall} \\
\cmidrule(lr){3-7} \cmidrule(lr){8-12}
Dataset & $n$ & RF & Prop. & WRF & KNE & KNU & RF & Prop. & WRF & KNE & KNU \\
\midrule
adult                & 4\,999 & .851 & .852 & .850 & .851 & .851 & .607 & .611 & .609 & .607 & .608 \\
australian           &    690 & .872 & .872 & .872 & .870 & .872 & .865 & .868 & .871 & .865 & .867 \\
bank-marketing       & 4\,999 & .897 & .897 & .897 & .896 & .897 & .309 & .310 & .310 & .311 & .310 \\
banknote             & 1\,372 & .992 & .994 & .992 & .995 & .993 & .993 & .995 & .992 & .997 & .994 \\
breast-cancer-uci    &    286 & .715 & .717 & .715 & .696 & .714 & .360 & .369 & .361 & .376 & .368 \\
breast-w             &    683 & .971 & .972 & .971 & .970 & .971 & .966 & .967 & .965 & .963 & .965 \\
default-credit-card  & 5\,000 & .812 & .815 & .812 & .809 & .811 & .366 & .385 & .368 & .361 & .368 \\
diabetes             &    768 & .760 & .761 & .759 & .754 & .758 & .579 & .584 & .581 & .582 & .580 \\
eeg-eye-state        & 4\,000 & .869 & .881 & .869 & .891 & .873 & .823 & .842 & .825 & .854 & .831 \\
electricity          & 4\,999 & .836 & .836 & .836 & .837 & .836 & .761 & .763 & .764 & .770 & .764 \\
german-credit        & 1\,000 & .754 & .754 & .754 & .752 & .754 & .374 & .373 & .373 & .392 & .376 \\
haberman             &    306 & .703 & .704 & .700 & .672 & .700 & .282 & .289 & .282 & .292 & .283 \\
heart-disease        &    214 & .793 & .797 & .794 & .788 & .795 & .317 & .331 & .323 & .340 & .325 \\
heart-statlog        &    270 & .830 & .831 & .831 & .821 & .830 & .760 & .760 & .762 & .752 & .762 \\
hepatitis            &     83 & .879 & .881 & .852 & .879 & .883 & .493 & .500 & .513 & .587 & .520 \\
ionosphere           &    351 & .933 & .932 & .935 & .937 & .934 & .878 & .877 & .878 & .876 & .878 \\
jm1                  & 10\,885 & .816 & .817 & .816 & .809 & .816 & .226 & .241 & .228 & .237 & .227 \\
kr-vs-kp             & 3\,196 & .990 & .990 & .990 & .991 & .990 & .986 & .986 & .986 & .987 & .986 \\
magic-gamma          & 4\,000 & .863 & .863 & .863 & .861 & .862 & .725 & .727 & .726 & .724 & .725 \\
mammographic-mass    &    961 & .787 & .815 & .787 & .765 & .781 & .757 & .792 & .758 & .727 & .730 \\
musk                 & 6\,598 & 1.000 & 1.000 & 1.000 & 1.000 & 1.000 & .999 & 1.000 & 1.000 & 1.000 & .999 \\
nomao                & 4\,999 & .949 & .950 & .950 & .950 & .950 & .899 & .898 & .898 & .902 & .899 \\
oil                  &    937 & .965 & .965 & .966 & .965 & .965 & .228 & .239 & .244 & .247 & .236 \\
parkinsons           &    195 & .903 & .907 & .908 & .910 & .904 & .702 & .716 & .722 & .727 & .704 \\
phoneme              & 5\,404 & .905 & .905 & .905 & .908 & .906 & .825 & .825 & .826 & .828 & .827 \\
sonar                &    208 & .812 & .815 & .812 & .850 & .823 & .726 & .729 & .728 & .764 & .740 \\
spambase             & 4\,601 & .954 & .955 & .954 & .953 & .954 & .927 & .929 & .928 & .928 & .927 \\
spect                &    267 & .826 & .826 & .828 & .810 & .821 & .484 & .484 & .478 & .492 & .527 \\
thoracic-surgery     &    470 & .849 & .848 & .847 & .839 & .848 & .019 & .021 & .019 & .033 & .019 \\
tic-tac-toe          &    958 & .938 & .948 & .940 & .947 & .939 & .827 & .850 & .829 & .857 & .829 \\
transfusion          &    748 & .745 & .757 & .744 & .712 & .746 & .322 & .347 & .323 & .337 & .304 \\
wdbc                 &    569 & .957 & .960 & .958 & .959 & .958 & .934 & .938 & .933 & .937 & .934 \\
wine                 &    130 & .986 & .986 & .970 & .984 & .986 & .991 & .991 & .980 & .987 & .991 \\
wine-quality-red     & 1\,599 & .807 & .807 & .806 & .804 & .806 & .796 & .797 & .794 & .795 & .794 \\
wine-quality-white   & 4\,898 & .831 & .831 & .831 & .829 & .831 & .686 & .687 & .685 & .686 & .685 \\
yeast-me2            & 1\,484 & .967 & .967 & .967 & .966 & .967 & .109 & .113 & .116 & .144 & .118 \\
\bottomrule
\end{tabular}
\end{table}

\subsection{Weight Amplification (Full Table)}

\Cref{tab:amp-sweep-full} reports the accuracy gain ($\Delta$acc vs RF) of the proposed method under weight amplification $\alpha = 1 + K \cdot M \cdot S$ with $K \in \{0, 10, 20, 30\}$.
$K = 0$ corresponds to the unamplified default ($\alpha = 1$).
$K^{*}$ is the per-dataset CV-selected amplification factor (selected within the same 5-fold CV that estimates the weight table; see \S\ref{sec:amplification}).
The RF column gives RF's mean accuracy; all other entries are mean $\Delta$acc over 30 stratified repeats (70/30 train/test split).

\begin{table}[H]
\centering\scriptsize
\caption{Weight amplification: $\Delta\text{acc}$ vs RF for the proposed method at four fixed $K$ values and CV-selected $K^{*}$.
The rightmost column reports how often each $K \in \{0,10,20,30\}$ was chosen across 30 repeats.
Only datasets where $K^{*} \ne 0$ on at least 5 repeats are shown individually; aggregate rows cover all 36 datasets.}
\label{tab:amp-sweep-full}
\setlength{\tabcolsep}{4pt}
\begin{tabular}{@{}l r rrrrr l@{}}
\toprule
& & \multicolumn{5}{c}{$\Delta$ acc vs RF} & \\
\cmidrule(lr){3-7}
Dataset & RF & $K{=}0$ & $K{=}10$ & $K{=}20$ & $K{=}30$ & $K^{*}$ & $K^{*}$ chosen \\
\midrule
mammographic-mass   & .787 & +.027 & +.033 & +.038 & +.040 & +.039 & 1/2/8/19 \\
eeg-eye-state       & .869 & +.012 & +.016 & +.017 & +.018 & +.018 & 0/1/6/23 \\
tic-tac-toe         & .938 & +.010 & +.013 & +.016 & +.016 & +.016 & 6/3/6/15 \\
transfusion         & .745 & +.013 & +.014 & +.017 & +.019 & +.018 & 3/1/9/17 \\
heart-statlog       & .830 & +.002 & +.004 & +.006 & +.007 & +.005 & 15/5/6/4 \\
haberman            & .703 & +.001 & +.001 & +.003 & +.004 & +.003 & 12/3/8/7 \\
default-credit-card & .812 & +.003 & +.004 & +.004 & +.005 & +.005 & 3/3/6/18 \\
australian          & .872 & +.000 & +.001 & +.001 & $-$.001 & +.001 & 12/3/10/5 \\
adult               & .851 & +.001 & +.001 & +.001 & +.001 & +.001 & 4/5/7/14 \\
sonar               & .812 & +.003 & +.002 & +.004 & +.005 & +.003 & 18/3/4/5 \\
\midrule
\multicolumn{2}{@{}l}{All 36 (mean $\Delta$)}  & +.0026 & +.0031 & +.0037 & +.0038 & +.0036 \\
\multicolumn{2}{@{}l}{W / T / L vs RF}         & 29/2/5 & 30/2/4 & 32/2/2 & 29/2/5 & 29/2/5 \\
\multicolumn{2}{@{}l}{Min.\ regr ($>0.2$\,pp)} & 0/36   & 0/36   & 0/36   & 1/36   & 0/36 \\
\multicolumn{2}{@{}l}{Maj.\ regr ($>0.2$\,pp)} & 1/36   & 2/36   & 1/36   & 2/36   & 1/36 \\
\bottomrule
\end{tabular}
\end{table}

\section{Code and Data Availability}
\label{app:code}
\ifanon
The full implementation, all 36 preprocessed benchmark datasets, and scripts to reproduce every table and figure will be released publicly upon acceptance; an anonymised copy is provided with the submission.
\else
The full implementation, all 36 preprocessed benchmark datasets, and scripts to reproduce every table and figure are available at \url{https://github.com/DavidParkYJ/dwarfp}.
\fi
All methods share the same train/test split and the same trained forest within each repeat (seed $=42+\text{repeat}$); the proposed method's weight table is estimated by 5-fold cross-validation on the training fold only, with test data never used during weight estimation.

% ── Declarations ──────────────────────────────────────────────────────
\section*{Declarations}

\textbf{Funding.} This research received no external funding.

\textbf{Competing interests.} The author declares no competing interests.

\textbf{Data availability.} All 36 benchmark datasets are publicly available from the UCI \citep{dua2017} and OpenML \citep{vanschoren2014} repositories; preprocessed datasets and code to reproduce all results are available at \url{https://github.com/DavidParkYJ/dwarfp}.

\textbf{Ethics approval.} Not applicable.

% ── References ────────────────────────────────────────────────────────
\clearpage
\bibliographystyle{plainnat}

\end{document}